\documentclass[10pt,twocolumn,letterpaper]{article}
\ifdefined\XeTeXversion
  \usepackage[OT1]{fontenc}
\fi

\usepackage{content/cvpr}              

\usepackage{xspace}
\usepackage{placeins}
\usepackage{float}
\newcommand{\method}{SIEVE\xspace}

\DeclareMathOperator*{\argmax}{arg\,max}

\DeclareMathOperator{\avgpool}{AvgPool}

\usepackage{algorithm}
\usepackage{algpseudocode}
\usepackage{makecell}
\definecolor{cvprblue}{rgb}{0.21,0.49,0.74}
\usepackage{colortbl}                       
\definecolor{ourshade}{rgb}{0.93,0.955,0.99}
\definecolor{avgshade}{rgb}{0.945,0.945,0.945}  
\usepackage{xcolor}
\usepackage{pifont}
\usepackage{soul}
\definecolor{improve}{HTML}{147A57}
\definecolor{regress}{HTML}{C8102E}
\definecolor{bbllava}{HTML}{DCE9F7}
\definecolor{bbqwenv2}{HTML}{E1F0DD}
\definecolor{bbqwenv3}{HTML}{FBE9CF}
\definecolor{bbintern}{HTML}{EBDDF3}
\definecolor{sectshade}{HTML}{E9ECF1}
\definecolor{supershade}{HTML}{CDD6E4}
\newcommand{\cmark}{\ding{51}}%
\newcommand{\xmark}{\ding{55}}
\newcommand{\yes}{\cellcolor{improve!12}\textcolor{improve}{\cmark}}
\newcommand{\no}{\cellcolor{regress!9}\textcolor{regress}{\xmark}}

\usepackage[pagebackref,breaklinks,colorlinks,allcolors=cvprblue]{hyperref}
\AtBeginDocument{\crefname{table}{Table}{Tables}}   
\definecolor{improve}{HTML}{147A57}

\newcommand{\dbad}[1]{\textcolor{red}{\boldmath$#1$}}

\def\paperID{26928} 
\def\confName{CVPR}
\def\confYear{2027}

\title{SIEVE: Preserving Cross-Image Evidence via Depth-Aware Token Compression}

\author{Nitish Shukla\textsuperscript{1} \quad Surgan Jandial\textsuperscript{2,*,\dag} \quad
 Nhat-Tan Bui\textsuperscript{3,*} \quad Arun Ross\textsuperscript{1}\\[2pt]
\textsuperscript{1}Michigan State University \quad
\textsuperscript{2}DatologyAI \quad
\textsuperscript{3}Carnegie Mellon University\\[2pt]
\normalsize
\textsuperscript{*}Equal contribution \quad
Corresponding author: {\tt\small shuklan3@msu.edu}\\[2pt]
{\hypersetup{urlcolor=red}\href{https://github.com/nitishshukla86/SIEVE}%
{\tt\small https://nitishshukla86.github.io/SIEVE/}}
}
\DeclareRobustCommand{\added}[1]{#1} 
\usepackage{xcolor}

\begin{document}
\twocolumn[{%
\renewcommand\twocolumn[1][]{#1}%
\maketitle

\begin{center}
  \captionsetup{type=figure}
  \includegraphics[width=\textwidth]{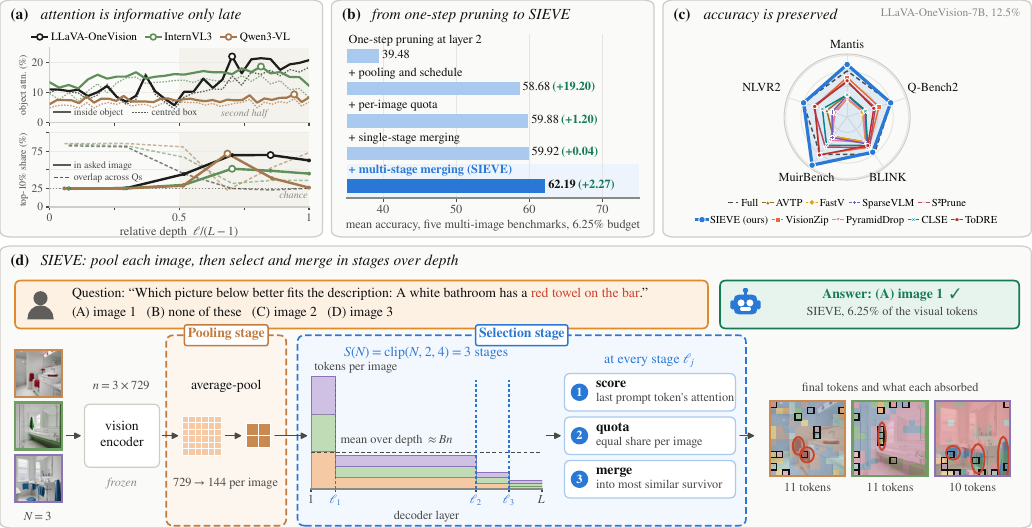}
  \vspace{-3.5mm}

\captionof{figure}{\textbf{Overview of \method.}
\textbf{(a, upper)} Attention to queried objects (solid) versus same-area central boxes (dotted) on COCO~\cite{Lin_2014}; query-relevant attention concentrates in later decoder layers across backbones.
\textbf{(a, lower)} Queried-image share of global top-$10\%$ tokens (solid) and cross-question overlap (dashed) across decoder depth, motivating per-image allocation and deferred selection.
\textbf{(b)} Component contributions at a $6.25\%$ budget, averaged over five benchmarks; progressive merging adds $2.27$ points.
\textbf{(c)} At a $12.5\%$ budget, \method leads on five multi-image benchmarks.
\textbf{(d)} \method progressively selects equal per-image quotas and merges discarded tokens within each image. Colors denote the absorbing tokens.}
\label{fig:teaser}
\end{center}
}]

\iftoggle{cvprfinal}{%
{\renewcommand{\thefootnote}{\dag}%
\footnotetext{Work done at Carnegie Mellon University.}}}{}
\begin{abstract}
Multi-image vision-language models (VLMs) incur substantial computational costs when processing long visual sequences. Compressing these sequences must preserve complementary evidence across images and keep tokens from different images separate. Global selection can underrepresent question-relevant images, and pruning in early decoder layers risks removing relevant content before sufficient multimodal processing, yet existing methods spend most of the budget on these early layers. We propose SIEVE, a training-free, single-pass framework that jointly allocates visual tokens across images and decoder depth. SIEVE first pools each image
into a compact spatial grid to reduce early processing cost. It then progressively selects representatives using prompt attention under equal per-image quotas and
aggregates unselected tokens into similar representatives from the same image. These operations keep every image represented and its tokens separate from other images,
while refining the retained representations. A closed-form schedule sets retention counts to target a prescribed average token budget across decoder layers. Across four backbones, SIEVE achieves the highest average performance at a 6.25\% budget across five multi-image, four single-image, two video, and five embodied application benchmarks. On LLaVA-OneVision-7B, it outperforms the strongest baseline by $9.96$ and $9.37$ points at $6.25\%$ and $12.5\%$ budgets.
\end{abstract}

\section{Introduction}
\label{sec:intro}
Multi-image VLMs integrate visual evidence across inputs~\cite{jiang2024mantisinterleavedmultiimageinstruction,
wang2024muirbenchcomprehensivebenchmarkrobust}. Their computational cost, however, grows with the number of images, as each image contributes additional visual tokens. Token compression reduces this cost, but multi-image inputs raise a specific challenge.
Task-relevant evidence can be complementary across images, so keeping more tokens from one image cannot compensate for removing the evidence needed from another. Effective
compression must therefore preserve complementary visual evidence while controlling the cost of processing it.

Prior work reduces visual tokens before the decoder~\cite{Yang_2025_CVPR,Alvar_2025_CVPR},
within the decoder~\cite{Chen_2024,Xing_2025_CVPR,pmlr-v267-zhang25s}, or at both stages~\cite{li2026todre}, and recent multi-image methods adapt retention to estimated image importance~\cite{zhang2026multiimagevisualtokenpruning}. We instead treat aggressive multi-image compression as a \emph{joint token allocation problem across images and decoder depth}. The retained tokens must support cross-image reasoning throughout inference while
staying within a limited computational budget.

This requires three key properties. \textbf{(1) Cross-image coverage.} A global ranking can leave too few tokens for an image containing evidence the task requires (\cref{fig:teaser}(a), bottom). \textbf{(2) Image identity.} Similar tokens from different images can describe different observations. If two images each contain a car, merging by similarity alone may merge distinct car representations into one token, and the model may lose source attribution for each car and produce incorrect counts across the images. \textbf{(3) Evidence across depth.} Early reduction can remove useful tokens before their relevance is clear, whereas delaying reduction keeps costly tokens from every image in the early layers.

\Cref{fig:teaser}(a) highlights two coupled challenges: when to select tokens and how to allocate the token budget across images. In the latter half of the decoder, attention from the final prompt token to the queried object peaks, and the queried image accounts for $51$--$72\%$ of globally top-ranked visual tokens. In contrast, early-layer attention is unreliable for global token ranking. At layer~2, the global top-$10\%$ tokens overlap by $82$--$85\%$ across questions targeting different images, allocate only the uniform $25\%$ share to the queried image, and exhibit strong positional bias (e.g., $49\%$ on the last image for LLaVA-OneVision-7B \cite{li2024llavaonevisioneasyvisualtask} and $63\%$ on the first for Qwen3-VL-8B \cite{bai2025qwen3vltechnicalreport}). Yet compression cannot be deferred: retaining all visual tokens through layer~2 already consumes $171\%$ of a $6.25\%$ depth-averaged budget on LLaVA-OneVision-7B (\cref{eq:prefix}). These observations motivate per-image pooling and equal quotas to reduce early costs and preserve coverage, with attention guiding within-image selection over decoder depth.

\Cref{fig:teaser}(b) progressively incorporates these components into one-step pruning. Pre-decoder pooling and budget-aware scheduling provide the largest gains by reducing early-layer costs. Equal per-image quotas prevent unreliable global rankings from underrepresenting individual images. Finally, similarity-based merging preserves information from unselected tokens, while repeated selection and merging progressively refine visual representations through image--text interaction.

\begin{table}[t]
  \centering
\caption{\textbf{Comparison with prior methods.} Unlike existing approaches, \method unifies pre-decoding compression, text-guided progressive token selection, balanced per-image retention, and within-image token aggregation in a single framework.}
  \label{tab:properties}
  \scriptsize
  \setlength{\tabcolsep}{2pt}
  \renewcommand{\arraystretch}{1.0}
  \resizebox{\columnwidth}{!}{%
  \begin{tabular}{@{}lccccc@{}}
    \toprule
    Method
    & \makecell{Pre-decoder\\compression}
    & \makecell{Text-guided\\reduction}
    & \makecell{Progressive\\selection}
    & \makecell{Per-image\\coverage}
    & \makecell{Same-image\\aggregation} \\
    \midrule
    VisionZip~\cite{Yang_2025_CVPR} & \yes & \no & \no & \yes$^{\S}$ & \yes$^{\S}$ \\
    FastV~\cite{Chen_2024} & \no & \yes & \no & \no & \no \\
    PyramidDrop~\cite{Xing_2025_CVPR} & \no & \yes & \yes & \no & \no \\
    SparseVLM~\cite{pmlr-v267-zhang25s} & \no & \yes & \yes & \no & \no \\
    AVTP~\cite{zhang2026multiimagevisualtokenpruning} & \no & \no & \yes & \no & \no \\
    CLSE~\cite{chen2026spectralevolutionguidedtokenpruning} & \no & \no & \no & \no & \no \\
    S$^2$Prune~\cite{jia2026s2prunespatiallystructuredvisual} & \no & \no & \no & \yes$^{\S}$ & \no \\
    ToDRE~\cite{li2026todre} & \yes & \yes & \no & \no & \no \\
    \midrule
    \textbf{\method} & \yes & \yes & \yes & \yes & \yes \\
    \bottomrule
  \end{tabular}%
  }
  \par\vspace{2pt}
  \begin{minipage}{\columnwidth}
    \scriptsize
    Coverage requires retaining tokens from every image at each stage; aggregation stays within each image. Check: explicit mechanism; cross: absent. \S: applied independently per image.
  \end{minipage}
  \vspace{-2em}
\end{table}

We therefore propose \method, a training-free, single-pass framework (\cref{fig:teaser}(d)). Spatial pooling first produces a compact grid for each image. Within
the decoder, equal per-image quotas maintain coverage, prompt attention selects representatives, and each unselected token is merged into its most similar representative from the same image, which preserves image identity. An early selection stage limits
processing cost, and later stages refine the representatives at greater depths. A closed-form
schedule sets the number of retained tokens from the target budget, the stage locations, and
the retention decay. \method improves over pooling
alone by $3.97$ points averaged across four backbones, while applying FastV's pruning or PyramidDrop's schedule to the pooled
input remains $3.69$ or $3.16$ points below \method, respectively (\cref{sec:ablations}). \Cref{tab:properties} contrasts these
mechanisms with those of prior methods.

We evaluate \method on four VLM backbones with five multi-image, four single-image, two
video, and five embodied application benchmarks. At a 6.25\% budget, it achieves the highest multi-image average on all four backbones, as well as the highest single-image, video, and embodied application means across backbones (\cref{tab:multi,tab:transfer}). 
On LLaVA-OneVision-7B \cite{li2024llavaonevisioneasyvisualtask}, at a 6.25\% token budget, it retains 98.5\% of the uncompressed multi-image average and outperforms the strongest baseline by 9.96 points (\cref{tab:multi}). Our contributions are as follows:
\begin{itemize}
  \item We analyze cross-image coverage, image identity, and compression timing, revealing why token allocation must account for both images and decoder depth. We also derive a feasibility condition for delayed reduction under a depth-averaged token budget.
  \item We introduce \method, a training-free framework that combines spatial pooling, per-image representative selection, and progressive same-image aggregation with a closed-form retention schedule.
  \item We provide a comprehensive evaluation under aggressive compression, demonstrating broad applicability and measuring the contribution of each component.
\end{itemize}
\section{Related Work}
\label{sec:related}

\noindent\textbf{Visual token reduction.}
Several methods reduce visual tokens before the decoder. VisionZip~\cite{Yang_2025_CVPR} uses
encoder attention to select tokens and aggregates the remaining ones. 
CDPruner~\cite{NEURIPS2025_2433fec2} conditions diversity-based selection
on the instruction. 
Others apply token reduction within the decoder. FastV~\cite{Chen_2024}
prunes after an early layer, PyramidDrop~\cite{Xing_2025_CVPR} applies staged pruning,
SparseVLM~\cite{pmlr-v267-zhang25s} uses text raters,
and ToDRE~\cite{li2026todre} combines reduction before and within the decoder.
CLSE~\cite{chen2026spectralevolutionguidedtokenpruning} scores tokens by the change in their
high-frequency spatial information across adjacent decoder layers. S$^2$Prune~\cite{jia2026s2prunespatiallystructuredvisual} splits each image into regions, keeps at
least one token per region, then gives more tokens to regions with more structure. 
These in-decoder methods retain all tokens through the initial layers, consuming a
substantial fraction of a restrictive depth-averaged budget. Most of them also score tokens by early
attention, which our analysis identifies as a weak relevance signal. 
\method instead preserves coarse spatial representations of all images through pooling before the decoder, selects tokens at multiple
decoder depths, and merges unselected tokens only into representatives from the same image.

\noindent\textbf{Multi-image token reduction.}
Some methods are designed explicitly for multi-image inputs. AVAM~\cite{Zeng_2025_ICCV} selects
regions and decodes original and cropped inputs, requiring additional decoding.
ERA~\cite{wang2026eraentropyguidedvisualtoken} uses entropy-guided selection and token recycling.
AVTP~\cite{zhang2026multiimagevisualtokenpruning} selects the pruning layer and assigns
image-dependent retention ratios using hidden-state importance scores. Allocations based on
importance scores can be unreliable at early layers, and these methods do not explicitly control
token counts across decoder depth under a strict depth-averaged budget. \method instead assigns an
equal quota to each image at every stage to maintain image coverage, restricts aggregation to tokens from the same image, and uses a schedule to match the depth-averaged budget.
\section{Method}
\label{sec:method}

\subsection{Notation}
\label{sec:notation}

\noindent\textbf{Inputs and visual tokens.}
Let $I_1,\dots,I_N$ be the $N$ input images and $\mathbf{p}$ the text prompt. A frozen vision encoder $E$ and projector $g$ map each image to a grid of visual tokens
$X_i = g(E(I_i)) \in \mathbb{R}^{P_i\times d}$. The total number of visual tokens is $n = \sum_{i=1}^{N} P_i$. 

\noindent\textbf{Prompt sequence.}
Visual tokens are interleaved with the text embeddings to form the prefill sequence
$H^{(0)} \in \mathbb{R}^{T\times d}$, and $\mathbf{h}_t$ denotes row $t$ of the current $H^{(\ell)}$.
An \emph{owner map} $\omega : \{1,\dots,T\} \to \{0,1,\dots,N\}$ records the image of each
position, with $\omega(t)=i>0$ for a visual token of image $I_i$ and $\omega(t)=0$ for a text token.

\noindent\textbf{Decoder and surviving sets.} The language model has $L$ decoder layers $f_0,\dots,f_{L-1}$ with
$H^{(\ell+1)} = f_\ell\big(H^{(\ell)}\big)$, where $L$ depends on the backbone (\cref{sec:setup}).
A stage after layer $\ell$ acts on $H^{(\ell+1)}$. We write $\mathcal{V}^{(\ell)}$ for the visual
tokens that enter $f_\ell$, so $\mathcal{V}^{(0)}$ is the pooled set, and $\mathcal{V}^{(\ell)}_i = \{t \in \mathcal{V}^{(\ell)} \mid \omega(t) = i\}$ for the tokens of image $i$. With $n_\ell = |\mathcal{V}^{(\ell)}|$, we have $n \ge n_0 \ge n_1 \ge \dots \ge n_{L-1}$. Text tokens are not counted.

\noindent\textbf{Compression budget.}
The prefill cost of layer $\ell$ grows with the number of tokens it processes, so we
measure compression by the \emph{depth-averaged} visual token count
\begin{equation}
  \bar{n} \;=\; \frac{1}{L}\sum_{\ell=0}^{L-1} n_\ell \;=\; B\, n ,
  \qquad B \in (0,1] 
  \label{eq:budget}
\end{equation}

where $B$ is the \emph{token budget}. Under a fixed depth-averaged budget, retaining more tokens in early layers requires more aggressive pruning in later layers.

\noindent\textbf{Overview.}
\label{sec:overview}
\Cref{fig:teaser}(d) summarizes the two-phase pipeline. The first pools each image before the decoder (\cref{sec:pooling}). The second runs $S$ stages inside the decoder (\cref{sec:selection}). Each of these $S$ stages selects tokens with prompt attention under equal per-image quotas and merges every unselected token into its most similar kept token from the same image. A closed-form schedule determines the retained token count per stage, to target \cref{eq:budget}. \method is a training-free, single-pass framework without auxiliary models.

\subsection{Pooling stage}
\label{sec:pooling}

Because \cref{eq:budget} averages over all $L$ layers, tokens that survive until the
first reduction layer $\ell_1$ consume $n_0(\ell_1+1)$ of the total budget $B n L$.
Retaining all $n$ tokens up to layer $\ell_1$ therefore consumes
\begin{equation}
  \frac{n(\ell_1+1)}{B n L} \;=\; \frac{\ell_1+1}{B L}
  \label{eq:prefix}
\end{equation}

of the budget. This fraction grows as $B$ decreases and exceeds $1$ when $BL < \ell_1+1$.
In that case, no later pruning can meet the budget. We therefore spatially compress each image before the decoder. The pooled count must be small
enough to keep this prefix cost low, yet larger than the budget $Bn$, to retain candidates for subsequent selection. We set the pooled total to
$m=\lfloor\min(\alpha B,\beta)n\rfloor$, with $\alpha>1$ and $B<\beta<1$,
where the headroom $\alpha$ keeps $m$ above the budget, leaving additional
candidates for prompt-guided selection. The cap $\beta$ binds only when $\alpha B > \beta$, which no reported budget reaches (at $B=\tfrac14$, $\alpha B = \beta$), so it is not a tuned hyperparameter. Image $i$ receives 
\added{$m_i = \operatorname{round}(m P_i / n)$ tokens.} Its grid is adaptively average-pooled to approximately $m_i$
tokens, each placed at the center of its pooling cell, which preserves the spatial layout of the
image. Because grid sides are integers, the realized total $\tilde{m}$ can differ
slightly from $m$.

\subsection{Selection stage}
\label{sec:selection}

The first stage is applied early, after layer $\ell_1$, because every additional layer at full pooled width uses more of the budget and leaves fewer tokens for later stages. The other stages are placed in the second half of the remaining depth, where prompt attention best separates relevant tokens (\cref{fig:teaser}(a)), at layers
\begin{equation}
  \ell_j \;=\; \ell_1 + \Big\lfloor \tfrac{L-\ell_1}{2} \Big\rfloor
         + \Big\lfloor \tfrac{(j-1)(L-\ell_1)}{2S} \Big\rfloor ,
  \qquad j = 2,\dots,S.
  \label{eq:stages}
\end{equation}
Merging occurs after each $f_{\ell_j}$. The placement depends only on $\ell_1$, $S$ and $L$, with no per-benchmark tuning.

\noindent\textbf{Prompt-attention scores.} {At stage $j$, each visual token is scored by the attention it receives from the final prompt token. Let $A^{(j)}_{h,t}$ be this
attention weight to position $t$ in head $h$ of the $N_{\mathrm{h}}$ heads at layer $\ell_j$,
computed from the layer input $H^{(\ell_j)}$. We set}
\begin{equation}
  s_t \;=\; \frac{1}{N_{\mathrm{h}}}\sum_{h=1}^{N_{\mathrm{h}}}
  \frac{A^{(j)}_{h,t}}{\sum_{t' \in \mathcal{V}^{(\ell_j)}} A^{(j)}_{h,t'}} ,
  \qquad t \in \mathcal{V}^{(\ell_j)} .
  \label{eq:score}
\end{equation}
Renormalizing over visual tokens gives every head an equal weight.

\noindent\textbf{Per-image quotas.}
Global top-$k$ selection can allocate few or no tokens for some images, which may remove evidence needed for cross-image comparison. We therefore split the stage budget $k_j$ \emph{uniformly} across images,
$q_i=\mathrm{clip}(\lfloor k_j/N\rfloor,q_{\min},|\mathcal{V}^{(\ell_j)}_i|)$,
with a floor of $q_{\min}$ tokens per image and any remainder distributed greedily, so that
$\sum_i q_i = k_j$ whenever $N q_{\min} \le k_j \le n_{\ell_j}$ and $|\mathcal{V}^{(\ell_j)}_i|\ge q_{\min}$ for every image $i$. Each image then keeps its
$q_i$ highest scoring tokens, $\mathcal{K}_i \subseteq \mathcal{V}^{(\ell_j)}_i$, and
$\mathcal{K} = \bigcup_{i=1}^{N}\mathcal{K}_i$. The quota determines the retained token count per image, while attention scores determine their identities.

\noindent\textbf{Similarity merging.}
{Let $\mathcal{D} = \mathcal{V}^{(\ell_j)} \setminus \mathcal{K}$ denote the unselected tokens. Using the layer outputs $H^{(\ell_j+1)}$, each $t \in \mathcal{D}$ is assigned to its most similar kept token from the same image, $\pi(t) = \argmax_{u \in \mathcal{K}_{\omega(t)}} \cos(\mathbf{h}_t, \mathbf{h}_u)$, and merged into it by score-weighted averaging,}
\begin{equation}
  \mathbf{h}_u \;\leftarrow\;
  \frac{s_u\, \mathbf{h}_u \;+\; \sum_{t \,:\, \pi(t) = u} s_t\, \mathbf{h}_t}
       {s_u \;+\; \sum_{t \,:\, \pi(t) = u} s_t} ,
  \qquad u \in \mathcal{K} .
  \label{eq:merge}
\end{equation}
Similarity-based assignment follows the token-merging principle of aggregating related
representations~\cite{bolya2023tokenmergingvitfaster}. 
{Restricting $\pi$ to the same image preserves image identity, so similar tokens from different images, such as two cars in two images, are never fused. Removing this restriction lowers the four-backbone mean by $1.89$ points at a $6.25\%$ budget (\cref{sec:ablations}).}

{\noindent\textbf{Sequence update.}
The sequence entering layer $\ell_j+1$ keeps all text tokens and $\mathcal{K}$, so
$\mathcal{V}^{(\ell_j+1)} = \mathcal{K}$. Unselected tokens add no key-value entries to later layers.}

\noindent\textbf{Budget schedule.}
\label{sec:schedule}
{The kept counts follow a geometric schedule, $k_j = k_1 \gamma^{\,j-1}$ with
$\gamma \in (0,1)$. The first $\ell_1+1$ layers process the pooled tokens, and stage $j$
covers the $\delta_j = \ell_{j+1}-\ell_j$ layers after it, with $\ell_{S+1} := L-1$.
Substituting into \cref{eq:budget} gives}
\begin{equation}
  k_1 \;=\;
  \frac{B\, n\, L \;-\; m\,(\ell_1 + 1)}
       {\sum_{j=1}^{S} \gamma^{\,j-1} \delta_j}.
  \label{eq:solve}
\end{equation}
{Each $k_j$ is then rounded and clipped to $[1, m]$. In all our settings the smallest $k_S$
exceeds $N q_{\min}$ by at least $7$ tokens, so the scheduled counts always satisfy the quota floor.

If the numerator of \cref{eq:solve} is not positive, the first $\ell_1+1$ layers already use the whole budget.
Pooling keeps the numerator positive at every evaluated budget. Since the schedule uses $m$ rather than the realized $\tilde{m}$, the depth average departs slightly from $Bn$; all reported FLOPs and token fractions use the measured average.}

\noindent\textbf{Position indices.}
{Pooled tokens are numbered from zero in raster order of the pooled grid. Inside the decoder,
each retained token preserves its index, including gaps, and merging changes only its content.}

\noindent\textbf{Adaptive stage count.}
\label{sec:adaptive}
{Additional stages recompute selection and merge assignments within each image using deeper hidden states, but they also add cost and repeated merging. We therefore set the number of stages from the image count,}
\begin{equation}
  S(N) \;=\; \mathrm{clip}\big(N,\; S_{\min},\; S_{\max}\big), \quad S_{\min} = 2,\; S_{\max} = 4.
  \label{eq:adaptiveS}
\end{equation}
A prompt with one or two images uses two stages, while one with four or more uses four. The floor of two ensures that single-image prompts are also reduced in more than one stage; with a floor of one they would be reduced only once, after layer~$\ell_1$. The cap limits the extra cost of scoring and merging. Stage locations and token counts then follow from \cref{eq:stages,eq:solve} with $S = S(N)$.


\section{Experiments}
\label{sec:experiments}
\subsection{Setup}
\label{sec:setup}
Additional details about the setup are in the supplementary.

\noindent\textbf{Models and implementation.}
We evaluate \method on LLaVA-OneVision-7B~\cite{li2024llavaonevisioneasyvisualtask},
Qwen2.5-VL-7B~\cite{bai2025qwen25vltechnicalreport}, Qwen3-VL-8B~\cite{bai2025qwen3vltechnicalreport},
and InternVL3-8B~\cite{zhu2025internvl3exploringadvancedtraining}.
\method modifies only prefilling; decoding is greedy throughout. Qwen3-VL has $36$ decoder layers, and the other backbones have $28$. LLaVA-OneVision represents each image in a multi-image prompt with a $384\times384$ view
and $729$ visual tokens; Qwen2.5-VL uses dynamic resolution, so token counts depend on the input.

\noindent\textbf{Hyperparameters.}
We use one configuration for all backbones, budgets, and benchmarks, with no benchmark-specific tuning:
first merging layer $\ell_1=2$, maximum stage count $S_{\max}=4$, retention decay $\gamma=\tfrac12$,
per-image quota floor $q_{\min}=4$, pooling headroom $\alpha=3$, and pooling cap $\beta=0.75$ (neither the cap nor the quota floor is binding in the evaluated configurations).

\noindent\textbf{Benchmarks and baselines.}
We test on five multi-image tasks (Mantis-Eval~\cite{jiang2024mantisinterleavedmultiimageinstruction}, NLVR2~\cite{suhr-etal-2019-corpus}, MuirBench~\cite{wang2024muirbenchcomprehensivebenchmarkrobust}, BLINK~\cite{Fu_2024}, and Q-Bench2~\cite{Zhang_2024}), four single-image tasks (TextVQA~\cite{Singh_2019_CVPR}, ScienceQA-IMG~\cite{NEURIPS2022_11332b6b}, MMStar~\cite{NEURIPS2024_2f8ee6a3}, and POPE~\cite{li-etal-2023-evaluating}), two video tasks (Video-MME~\cite{fu2025video} and EgoSchema~\cite{mangalam2023egoschema}), and five embodied application tasks (ERQA~\cite{geminiroboticsteam2025geminiroboticsbringingai}, MV-RoboBench~\cite{feng2026seeingviewsbenchmarkingspatial}, OpenEQA~\cite{Majumdar_2024_CVPR}, RoboSpatial~\cite{song2025robospatial}, and NuScenes-QA~\cite{qian2023nuscenes}). The baselines are VisionZip~\cite{Yang_2025_CVPR}, FastV~\cite{Chen_2024},
PyramidDrop~\cite{Xing_2025_CVPR}, SparseVLM~\cite{pmlr-v267-zhang25s},
CLSE~\cite{chen2026spectralevolutionguidedtokenpruning}, S$^2$Prune~\cite{jia2026s2prunespatiallystructuredvisual},
ToDRE~\cite{li2026todre}, and AVTP~\cite{zhang2026multiimagevisualtokenpruning}.
Video clips follow each backbone's native $32$-frame video pipeline, while embodied tasks provide temporal observations or camera views. For compression, each frame---or each temporal patch for the Qwen models---is treated as one observation and uses the same spatial pooling, per-image budget allocation, and progressive select-and-merge schedule. This configuration is shared across benchmarks without task-specific training, architectural changes, or hyperparameter tuning.

\noindent\textbf{Token budget.}
Comparisons use the same depth-averaged visual-token retention $B\in\{\tfrac1{16},\tfrac18,\tfrac14\}$
(\cref{eq:budget}).
A method that keeps the full visual sequence through layer~$2$ of a $28$-layer decoder already spends
$171\%$ of the $6.25\%$ budget before pruning (\cref{eq:prefix}), so FastV and PyramidDrop exceed
this target even when reduced to one token. 

\setcounter{dbltopnumber}{1}
\setcounter{topnumber}{1}
\begin{table*}[t]
  \centering
  \caption{\textbf{Performance comparison for multi-image benchmarks.} The best average in each block is \textbf{bold}; \textit{Full} is the model with no compression. A dash marks a method that cannot run at that budget, explained in supplementary C2. 
  $\ddagger$: exceeds the target budget even when retaining only one token after pruning.}
  \label{tab:multi}
  \scriptsize
  \setlength{\tabcolsep}{1.0pt}
  \resizebox{\textwidth}{!}{%
  \begin{tabular}{@{}ll*{4}{ccccc>{\columncolor{avgshade}}c}@{}}
    \toprule
    Method & & \multicolumn{6}{c}{\cellcolor{bbllava}\textbf{LLaVA-OneVision-7B}} & \multicolumn{6}{c}{\cellcolor{bbqwenv2}\textbf{Qwen2.5-VL-7B}} & \multicolumn{6}{c}{\cellcolor{bbqwenv3}\textbf{Qwen3-VL-8B}} & \multicolumn{6}{c}{\cellcolor{bbintern}\textbf{InternVL3-8B}} \\
    \cmidrule(lr){3-8} \cmidrule(lr){9-14} \cmidrule(lr){15-20} \cmidrule(lr){21-26}
     & Venue & Mant. & NLVR2 & Muir & BLINK & QB2 & \textbf{Avg} & Mant. & NLVR2 & Muir & BLINK & QB2 & \textbf{Avg} & Mant. & NLVR2 & Muir & BLINK & QB2 & \textbf{Avg} & Mant. & NLVR2 & Muir & BLINK & QB2 & \textbf{Avg} \\
    \midrule
    \textit{Full} & & \textit{60.83} & \textit{89.22} & \textit{41.96} & \textit{49.87} & \textit{73.70} & \textit{63.12} & \textit{75.12} & \textit{87.90} & \textit{58.58} & \textit{57.44} & \textit{75.20} & \textit{70.85} & \textit{66.82} & \textit{87.18} & \textit{50.31} & \textit{58.86} & \textit{76.40} & \textit{67.91} & \textit{68.20} & \textit{85.72} & \textit{54.04} & \textit{56.34} & \textit{77.10} & \textit{68.28} \\
    \midrule
    \rowcolor{sectshade} \multicolumn{26}{l}{\textbf{\textit{Keep 6.25\% of visual tokens}}} \\
    VisionZip~\cite{Yang_2025_CVPR} & \textcolor{black!55}{\texttt{CVPR'25}} & 53.00 & 65.48 & 39.77 & 41.87 & 56.50 & 51.32 & 69.12 & 70.78 & 45.58 & 49.03 & 66.80 & 60.26 & 54.38 & 62.26 & 45.27 & 44.98 & 69.70 & 55.32 & 63.13 & 66.01 & 40.31 & 44.13 & 71.90 & 57.10 \\
    FastV~\cite{Chen_2024}$^{\ddagger}$ & \textcolor{black!55}{\texttt{ECCV'24}} & 36.87 & 49.45 & 26.04 & 39.24 & 45.80 & 39.48 & 42.86 & 49.86 & 26.65 & 39.72 & 46.20 & 41.06 & 30.88 & 50.62 & 43.85 & 34.77 & 47.10 & 41.44 & 41.47 & 50.84 & 20.92 & 36.93 & 45.20 & 39.07 \\
    PyramidDrop~\cite{Xing_2025_CVPR}$^{\ddagger}$ & \textcolor{black!55}{\texttt{CVPR'25}} & 36.87 & 49.45 & 26.04 & 39.24 & 45.80 & 39.48 & 42.86 & 49.86 & 26.65 & 39.72 & 46.20 & 41.06 & 30.88 & 50.62 & 43.85 & 34.77 & 47.10 & 41.44 & 41.47 & 50.84 & 20.92 & 36.93 & 45.20 & 39.07 \\
    AVTP~\cite{zhang2026multiimagevisualtokenpruning} & \textcolor{black!55}{\texttt{arXiv'26}} & 38.25 & 50.77 & 25.58 & 38.61 & 45.40 & 39.72 & 58.06 & 56.61 & 39.96 & 39.87 & 52.90 & 49.48 & 34.56 & 52.50 & 44.12 & 34.35 & 48.20 & 42.75 & 52.07 & 54.53 & 31.04 & 36.35 & 50.80 & 44.96 \\
    CLSE~\cite{chen2026spectralevolutionguidedtokenpruning} & \textcolor{black!55}{\texttt{ECCV'26}} & -- & -- & -- & -- & -- & -- & -- & -- & -- & -- & -- & -- & 32.72 & 50.54 & 43.58 & 32.40 & 58.60 & 43.57 & -- & -- & -- & -- & -- & -- \\
    S$^2$Prune~\cite{jia2026s2prunespatiallystructuredvisual} & \textcolor{black!55}{\texttt{arXiv'26}} & 42.86 & 67.86 & 32.23 & 40.24 & 52.00 & 47.04 & 66.36 & 66.15 & 44.03 & 42.94 & 60.70 & 56.04 & 54.84 & 61.71 & 43.23 & 39.51 & 62.10 & 52.28 & 58.53 & 63.69 & 40.58 & 42.08 & 56.00 & 52.18 \\
    ToDRE~\cite{li2026todre} & \textcolor{black!55}{\texttt{ECCV'26}} & 52.07 & 72.92 & 41.88 & 41.56 & 52.70 & \underline{52.23} & 70.51 & 73.47 & 48.65 & 48.24 & 68.90 & \underline{61.95} & 59.91 & 65.15 & 47.23 & 41.29 & 67.80 & \underline{56.28} & 66.36 & 76.17 & 51.81 & 48.19 & 73.80 & \underline{63.27} \\
    \cmidrule(lr){2-26}
    \rowcolor{ourshade} \textbf{\method} & & 63.59 & 83.61 & 47.77 & 45.50 & 70.50 & \textbf{62.19} & 71.43 & 78.05 & 50.96 & 49.50 & 68.40 & \textbf{63.67} & 60.83 & 73.20 & 47.27 & 44.29 & 68.50 & \textbf{58.82} & 67.74 & 78.30 & 50.92 & 49.50 & 71.40 & \textbf{63.57} \\
    \midrule
    \rowcolor{sectshade} \multicolumn{26}{l}{\textbf{\textit{Keep 12.5\% of visual tokens}}} \\
    VisionZip~\cite{Yang_2025_CVPR} & \textcolor{black!55}{\texttt{CVPR'25}} & 55.76 & 75.14 & 42.19 & 44.13 & 61.60 & \underline{55.76} & 71.43 & 78.79 & 51.58 & 52.50 & 70.80 & 65.02 & 61.75 & 72.08 & 45.81 & 50.76 & 71.70 & \underline{60.42} & 67.74 & 74.11 & 50.54 & 48.92 & 76.30 & 63.52 \\
    FastV~\cite{Chen_2024} & \textcolor{black!55}{\texttt{ECCV'24}} & 38.71 & 52.05 & 29.85 & 40.08 & 45.10 & 41.16 & 50.23 & 53.24 & 31.85 & 42.19 & 59.80 & 47.46 & 47.00 & 55.72 & 44.65 & 45.55 & 67.90 & 52.17 & 54.38 & 61.12 & 31.38 & 41.61 & 66.70 & 51.04 \\
    PyramidDrop~\cite{Xing_2025_CVPR} & \textcolor{black!55}{\texttt{CVPR'25}} & 39.17 & 50.01 & 32.38 & 40.45 & 46.40 & 41.68 & 40.09 & 51.60 & 30.46 & 40.29 & 55.20 & 43.53 & 41.94 & 53.98 & 45.62 & 40.35 & 68.00 & 49.98 & 48.39 & 53.41 & 26.54 & 39.61 & 60.90 & 45.77 \\
    SparseVLM~\cite{pmlr-v267-zhang25s} & \textcolor{black!55}{\texttt{ICML'25}} & 41.47 & 53.93 & 30.69 & 41.45 & 48.00 & 43.11 & 57.60 & 60.10 & 34.31 & 45.08 & 64.70 & 52.36 & 63.59 & 67.36 & 45.62 & 49.55 & 72.00 & 59.62 & 56.68 & 68.68 & 32.65 & 46.50 & 70.20 & 54.94 \\
    AVTP~\cite{zhang2026multiimagevisualtokenpruning} & \textcolor{black!55}{\texttt{arXiv'26}} & 41.94 & 60.40 & 29.92 & 40.14 & 48.70 & 44.22 & 63.13 & 68.24 & 47.35 & 45.66 & 62.10 & 57.30 & 52.53 & 60.33 & 44.85 & 37.45 & 61.60 & 51.35 & 59.91 & 62.35 & 39.50 & 42.35 & 60.50 & 52.92 \\
    CLSE~\cite{chen2026spectralevolutionguidedtokenpruning} & \textcolor{black!55}{\texttt{ECCV'26}} & 41.94 & 65.87 & 35.92 & 42.82 & 47.90 & 46.89 & 70.97 & 71.84 & 47.23 & 48.71 & 68.10 & 61.37 & 59.91 & 65.61 & 43.31 & 46.03 & 73.30 & 57.63 & 61.29 & 72.83 & 40.12 & 49.76 & 71.30 & 59.06 \\
    S$^2$Prune~\cite{jia2026s2prunespatiallystructuredvisual} & \textcolor{black!55}{\texttt{arXiv'26}} & 52.07 & 77.02 & 37.08 & 44.45 & 58.00 & 53.72 & 69.59 & 78.18 & 52.31 & 47.87 & 67.80 & 63.15 & 59.91 & 72.50 & 45.77 & 43.24 & 67.80 & 57.84 & 63.59 & 73.68 & 46.00 & 46.82 & 66.30 & 59.28 \\
    ToDRE~\cite{li2026todre} & \textcolor{black!55}{\texttt{ECCV'26}} & 53.00 & 75.27 & 42.50 & 42.82 & 56.40 & 54.00 & 69.59 & 79.12 & 54.00 & 50.92 & 73.60 & \underline{65.45} & 63.59 & 72.61 & 47.88 & 44.24 & 71.90 & 60.04 & 71.89 & 80.51 & 54.00 & 51.97 & 75.30 & \textbf{66.73} \\
    \cmidrule(lr){2-26}
    \rowcolor{ourshade} \textbf{\method} & & 65.90 & 88.83 & 49.31 & 48.19 & 73.40 & \textbf{65.13} & 72.81 & 82.16 & 52.35 & 52.03 & 71.70 & \textbf{66.21} & 65.44 & 79.25 & 48.04 & 49.50 & 71.90 & \textbf{62.83} & 69.12 & 80.90 & 52.15 & 51.29 & 72.60 & \underline{65.21} \\
    \midrule
    \rowcolor{sectshade} \multicolumn{26}{l}{\textbf{\textit{Keep 25\% of visual tokens}}} \\
    VisionZip~\cite{Yang_2025_CVPR} & \textcolor{black!55}{\texttt{CVPR'25}} & 59.91 & 84.27 & 42.62 & 48.08 & 69.10 & \underline{60.80} & 71.43 & 84.00 & 55.04 & 55.18 & 73.60 & 67.85 & 65.90 & 81.23 & 47.88 & 55.39 & 73.50 & 64.78 & 70.05 & 79.89 & 54.31 & 52.08 & 76.20 & 66.51 \\
    FastV~\cite{Chen_2024} & \textcolor{black!55}{\texttt{ECCV'24}} & 50.23 & 70.26 & 36.19 & 43.45 & 58.30 & 51.69 & 68.20 & 72.20 & 47.73 & 52.45 & 70.70 & 62.26 & 63.59 & 73.17 & 47.27 & 53.02 & 73.40 & 62.09 & 67.74 & 82.47 & 51.23 & 53.76 & 74.50 & 65.94 \\
    PyramidDrop~\cite{Xing_2025_CVPR} & \textcolor{black!55}{\texttt{CVPR'25}} & 52.53 & 77.39 & 38.19 & 44.45 & 65.70 & 55.65 & 69.12 & 71.85 & 44.35 & 50.24 & 75.20 & 62.15 & 67.74 & 73.25 & 47.12 & 51.76 & 74.40 & 62.85 & 67.74 & 78.96 & 48.85 & 54.55 & 75.10 & 65.04 \\
    SparseVLM~\cite{pmlr-v267-zhang25s} & \textcolor{black!55}{\texttt{ICML'25}} & 52.53 & 77.31 & 35.27 & 44.35 & 64.80 & 54.85 & 72.81 & 82.63 & 46.58 & 53.97 & 73.90 & 65.98 & 67.74 & 82.19 & 48.85 & 56.23 & 75.70 & \textbf{66.14} & 65.90 & 83.03 & 49.54 & 55.60 & 74.80 & 65.77 \\
    AVTP~\cite{zhang2026multiimagevisualtokenpruning} & \textcolor{black!55}{\texttt{arXiv'26}} & 48.85 & 69.26 & 37.23 & 41.35 & 58.60 & 51.06 & 68.20 & 76.59 & 54.08 & 50.13 & 68.50 & 63.50 & 60.83 & 72.60 & 47.81 & 43.14 & 68.30 & 58.53 & 64.52 & 70.55 & 46.23 & 47.13 & 68.70 & 59.43 \\
    CLSE~\cite{chen2026spectralevolutionguidedtokenpruning} & \textcolor{black!55}{\texttt{ECCV'26}} & 55.30 & 78.94 & 39.69 & 43.87 & 66.00 & 56.76 & 70.97 & 82.13 & 54.77 & 53.13 & 72.90 & 66.78 & 65.44 & 79.59 & 47.73 & 52.03 & 74.00 & 63.76 & 66.82 & 82.37 & 50.04 & 53.29 & 75.70 & 65.64 \\
    S$^2$Prune~\cite{jia2026s2prunespatiallystructuredvisual} & \textcolor{black!55}{\texttt{arXiv'26}} & 56.22 & 84.20 & 42.46 & 45.24 & 66.00 & 58.82 & 73.73 & 84.68 & 56.00 & 52.76 & 72.40 & 67.91 & 62.21 & 80.72 & 48.27 & 51.18 & 70.90 & 62.66 & 66.36 & 80.95 & 49.58 & 50.71 & 71.60 & 63.84 \\
    ToDRE~\cite{li2026todre} & \textcolor{black!55}{\texttt{ECCV'26}} & 58.06 & 81.41 & 43.19 & 46.34 & 65.40 & 58.88 & 73.73 & 83.71 & 56.77 & 54.13 & 72.40 & \underline{68.15} & 64.52 & 79.86 & 48.50 & 51.39 & 75.70 & 63.99 & 69.59 & 83.13 & 55.85 & 53.87 & 76.10 & \underline{67.71} \\
    \cmidrule(lr){2-26}
    \rowcolor{ourshade} \textbf{\method} & & 68.20 & 90.41 & 47.27 & 51.24 & 76.10 & \textbf{66.64} & 73.27 & 84.48 & 55.85 & 54.81 & 72.90 & \textbf{68.26} & 66.36 & 83.36 & 47.88 & 54.02 & 74.10 & \underline{65.14} & 72.35 & 82.96 & 54.12 & 54.97 & 75.30 & \textbf{67.94} \\
    \bottomrule
  \end{tabular}%
  }
  \vspace{-3mm}
\end{table*}
\begin{table*}[t]
  \centering
  \caption{\textbf{Performance on video, embodied application, and single-image benchmarks.} Averaged across two benchmarks for video, five benchmarks for embodied application, and four benchmarks for single-image. \method achieves the highest mean across backbones in all three settings at both budgets. --: unavailable or infeasible. $\ddagger$: exceeds the target budget even when retaining only one token after pruning.}
  \label{tab:transfer}
  \label{tab:video}
  \label{tab:robotics}
  \label{tab:single-mean}
  \scriptsize
  \renewcommand{\arraystretch}{0.92}
  \setlength{\tabcolsep}{2.2pt}
  \resizebox{\textwidth}{!}{%
  \begin{tabular}{l*{3}{cccc>{\columncolor{avgshade}}r}}
    \toprule
    & \multicolumn{5}{c}{\cellcolor{sectshade}\textbf{Video}} & \multicolumn{5}{c}{\cellcolor{sectshade}\textbf{Embodied Application}} & \multicolumn{5}{c}{\cellcolor{sectshade}\textbf{Single-image}} \\
    \cmidrule(lr){2-6}\cmidrule(lr){7-11}\cmidrule(lr){12-16}
    Method & \cellcolor{bbllava}\textbf{OneVision} & \cellcolor{bbqwenv2}\textbf{Qwen2.5} & \cellcolor{bbqwenv3}\textbf{Qwen3} & \cellcolor{bbintern}\textbf{InternVL} & \textbf{Avg} & \cellcolor{bbllava}\textbf{OneVision} & \cellcolor{bbqwenv2}\textbf{Qwen2.5} & \cellcolor{bbqwenv3}\textbf{Qwen3} & \cellcolor{bbintern}\textbf{InternVL} & \textbf{Avg} & \cellcolor{bbllava}\textbf{OneVision} & \cellcolor{bbqwenv2}\textbf{Qwen2.5} & \cellcolor{bbqwenv3}\textbf{Qwen3} & \cellcolor{bbintern}\textbf{InternVL} & \textbf{Avg} \\
    \midrule
    \textit{Full} & \textit{60.52} & \textit{60.80} & \textit{66.43} & \textit{67.78} & \textit{63.88} & \textit{40.15} & \textit{39.10} & \textit{45.67} & \textit{42.21} & \textit{41.78} & \textit{80.67} & \textit{80.59} & \textit{81.76} & \textit{84.44} & \textit{81.87} \\
    \midrule
    \rowcolor{sectshade} \multicolumn{16}{l}{\textbf{\textit{Keep 6.25\% of visual tokens}}} \\
    VisionZip~\cite{Yang_2025_CVPR} & 52.39 & 53.86 & 57.48 & 55.23 & 54.74 & 37.31 & 35.39 & 38.27 & 36.03 & 36.75 & 65.14 & 62.04 & 56.24 & 56.78 & \underline{60.05} \\
    FastV~\cite{Chen_2024}$^{\ddagger}$ & 37.64 & 35.36 & 39.67 & 38.76 & 37.86 & 32.45 & 28.60 & 37.16 & 35.76 & 33.49 & 40.13 & 38.85 & 42.28 & 40.12 & 40.35 \\
    PyramidDrop~\cite{Xing_2025_CVPR}$^{\ddagger}$ & 37.64 & 35.36 & 39.75 & 38.76 & 37.88 & 32.46 & 28.65 & 37.14 & 35.75 & 33.50 & 40.13 & 38.85 & 42.28 & 40.12 & 40.35 \\
    AVTP~\cite{zhang2026multiimagevisualtokenpruning} & 39.00 & 48.52 & 46.86 & 46.08 & 45.12 & 31.87 & 29.70 & 37.46 & 33.67 & 33.18 & -- & -- & -- & -- & -- \\
    CLSE~\cite{chen2026spectralevolutionguidedtokenpruning} & -- & -- & 40.32 & -- & -- & -- & -- & 36.24 & -- & -- & -- & -- & 42.67 & -- & -- \\
    S$^2$Prune~\cite{jia2026s2prunespatiallystructuredvisual} & 46.73 & 49.26 & 53.73 & 49.57 & 49.82 & 34.20 & 33.04 & 40.08 & 34.95 & 35.57 & 56.63 & 54.91 & 50.30 & 59.18 & 55.25 \\
    ToDRE~\cite{li2026todre} & 57.41 & 57.81 & 59.32 & 61.36 & \underline{58.97} & 38.37 & 35.10 & 40.46 & 40.28 & \underline{38.55} & -- & -- & -- & -- & -- \\
    \rowcolor{ourshade} \textbf{\method} & 57.96 & 58.52 & 62.21 & 60.36 & \textbf{59.76} & 38.42 & 36.11 & 42.07 & 39.38 & \textbf{39.00} & 69.66 & 58.19 & 59.53 & 64.44 & \textbf{62.96} \\
    \midrule
    \rowcolor{sectshade} \multicolumn{16}{l}{\textbf{\textit{Keep 12.5\% of visual tokens}}} \\
    VisionZip~\cite{Yang_2025_CVPR} & 58.58 & 56.42 & 62.78 & 61.40 & 59.79 & 39.59 & 36.84 & 41.29 & 39.12 & 39.21 & 71.97 & 68.67 & 67.07 & 66.39 & \underline{68.52} \\
    FastV~\cite{Chen_2024} & 46.83 & 47.48 & 53.34 & 50.60 & 49.56 & 33.04 & 30.62 & 38.11 & 36.13 & 34.48 & 49.62 & 45.96 & 47.14 & 58.41 & 50.28 \\
    PyramidDrop~\cite{Xing_2025_CVPR} & 43.64 & 45.42 & 48.96 & 44.48 & 45.63 & 32.62 & 30.84 & 37.43 & 35.57 & 34.11 & 47.88 & 45.98 & 46.23 & 49.37 & 47.37 \\
    SparseVLM~\cite{pmlr-v267-zhang25s} & 48.83 & 55.20 & 60.51 & 56.68 & 55.31 & 32.81 & 31.25 & 40.29 & 38.84 & 35.80 & 52.82 & 56.52 & 58.84 & 65.47 & 58.41 \\
    AVTP~\cite{zhang2026multiimagevisualtokenpruning} & 45.27 & 54.19 & 54.08 & 52.55 & 51.52 & 32.64 & 32.32 & 37.49 & 34.29 & 34.19 & -- & -- & -- & -- & -- \\
    CLSE~\cite{chen2026spectralevolutionguidedtokenpruning} & 53.35 & 56.39 & 57.58 & 60.00 & 56.83 & 34.90 & 33.42 & 40.86 & 39.64 & 37.21 & 59.84 & 67.06 & 57.15 & 71.73 & 63.95 \\
    S$^2$Prune~\cite{jia2026s2prunespatiallystructuredvisual} & 56.25 & 56.08 & 60.16 & 57.88 & 57.59 & 37.90 & 36.14 & 41.84 & 38.98 & 38.71 & 68.84 & 66.27 & 58.86 & 68.98 & 65.74 \\
    ToDRE~\cite{li2026todre} & 58.47 & 60.50 & 61.13 & 62.70 & \underline{60.70} & 38.42 & 36.20 & 41.88 & 41.50 & \underline{39.50} & -- & -- & -- & -- & -- \\
    \rowcolor{ourshade} \textbf{\method} & 57.93 & 60.65 & 64.48 & 61.77 & \textbf{61.21} & 40.21 & 37.88 & 42.58 & 40.50 & \textbf{40.29} & 71.97 & 65.21 & 65.70 & 75.49 & \textbf{69.59} \\
    \bottomrule
  \end{tabular}%
  }
  \vspace{-3mm}
\end{table*}
\begin{table}[t]
  \centering
  \caption{\textbf{Efficiency analysis.} Average across five benchmarks. Storage denotes key-value cache size. Prefill includes any pre-LLM token reduction. Measurements use a single NVIDIA H100.}
  \label{tab:efficiency}
  \renewcommand{\arraystretch}{0.92}
  \resizebox{0.9\columnwidth}{!}{%
  \begin{tabular}{lccc>{\columncolor{avgshade}}c}
    \toprule
    Method
    & \cellcolor{bbllava}\textbf{FLOPs (T) $\downarrow$}
    & \cellcolor{bbllava}\textbf{Storage (MB) $\downarrow$}
    & \cellcolor{bbllava}\textbf{Prefill (ms) $\downarrow$}
    & \cellcolor{bbqwenv2}\textbf{Acc $\uparrow$} \\
    \midrule
    \textit{Full} & \textit{26.46} & \textit{107.4} & \textit{100.4} & \textit{63.12} \\
    \midrule
    \rowcolor{sectshade} \multicolumn{5}{l}{\textbf{\textit{Keep 12.5\% of visual tokens}}} \\
    VisionZip~\cite{Yang_2025_CVPR} & \underline{4.06} & \underline{16.9} & \underline{47.0} & \underline{55.76} \\
    FastV~\cite{Chen_2024} & 4.13 & \underline{16.9} & 47.9 & 41.16 \\
    PyramidDrop~\cite{Xing_2025_CVPR} & 4.13 & \underline{16.9} & 52.0 & 41.68 \\
    SparseVLM~\cite{pmlr-v267-zhang25s} & 4.13 & \underline{16.9} & 53.8 & 43.11 \\
    AVTP~\cite{zhang2026multiimagevisualtokenpruning} & 4.07 & \underline{16.9} & \textbf{46.6} & 44.22 \\
    CLSE~\cite{chen2026spectralevolutionguidedtokenpruning} & 4.10 & \underline{16.9} & 54.4 & 46.89 \\
    S$^2$Prune~\cite{jia2026s2prunespatiallystructuredvisual} & 4.10 & 17.0 & 59.7 & 53.72 \\
    ToDRE~\cite{li2026todre} & \underline{4.06} & \underline{16.9} & 71.3 & 54.00 \\
    \cmidrule(lr){2-5}
    \rowcolor{ourshade} \textbf{\method} & \textbf{4.02} & \textbf{16.7} & 53.5 & \textbf{65.13} \\
    \bottomrule
  \end{tabular}%
  }
  \vspace{-1mm}
\end{table}

\subsection{Multi-image results}
\label{sec:results-multi}

We first evaluate the setting targeted directly by \method: heterogeneous image sets in which task-relevant evidence may be distributed across multiple images.



\begin{figure}[t]
  \centering
  \includegraphics[width=\linewidth]{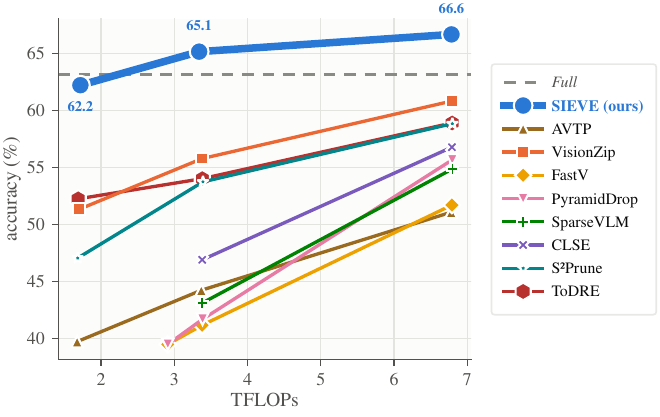}
  \vspace{-5mm}
  \caption{\textbf{Performance-efficiency trade-off curves} on LLaVA-OneVision at $B\in\{\tfrac1{16},\tfrac18,\tfrac14\}$ (five multi-image benchmark averages). \method has the highest accuracy at every compute level. x-axis: estimated visual-token prefill TFLOPs.}
  \label{fig:flops}
  \vspace{-4mm}
\end{figure}

\noindent\textbf{Performance under tight budgets.}
\Cref{tab:multi} shows that \method achieves the highest mean score on all four backbones at $B=\tfrac1{16}$, improving over the strongest baseline by $9.96$ points on LLaVA-OneVision and $1.72$--$2.54$ on Qwen2.5-VL and Qwen3-VL. \method retains $87$--$99\%$ of the uncompressed mean at $B=\tfrac1{16}$ and $93$--$103\%$ at $B=\tfrac18$.
On LLaVA-OneVision it exceeds the uncompressed model at $B=\tfrac18$ and $\tfrac14$ ($65.13$ and $66.64$ vs.\ $63.12$).
\Cref{fig:flops} shows that \method is the most accurate at every compute level.

\begin{figure}[t]
  \centering
  \includegraphics[width=0.9\linewidth]{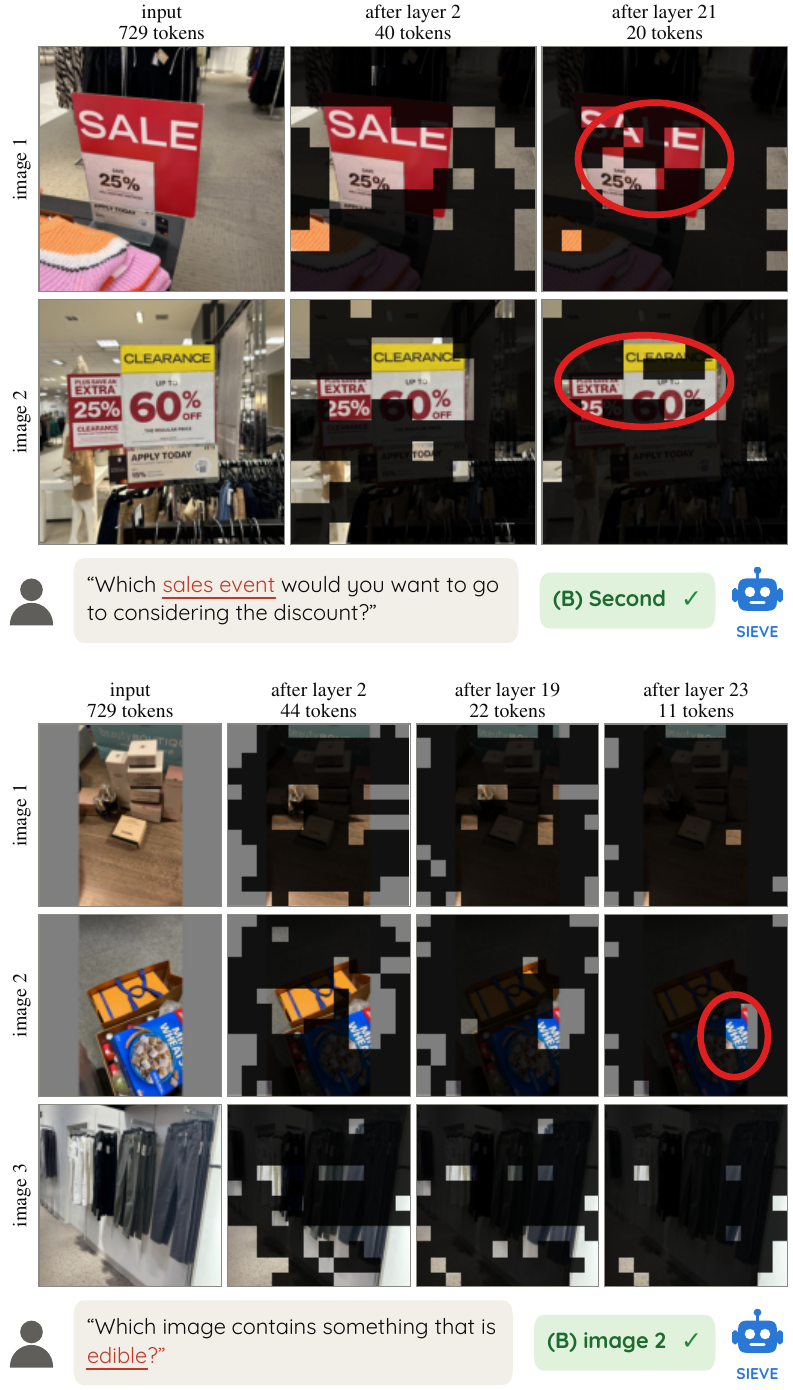}
  \caption{\textbf{Retained visual tokens at each merging stage}, LLaVA-OneVision-7B at $6.25\%$. Token counts are per image. Darkened cells are no longer separate tokens. Stages follow \cref{eq:stages,eq:adaptiveS}. \textcolor{red}{Red rings} mark the target objects (discount and cereal box).}
  \label{fig:qualitative}
  \vspace{-1mm}
\end{figure}

\noindent\textbf{Baseline performance under delayed compression.}
On LLaVA-OneVision at $B=\tfrac18$, FastV, PyramidDrop, and SparseVLM achieve $41.16$--$43.11$ points, trailing \method by $22.0$--$24.0$ points.
They keep all tokens through layer~$2$, which already uses $86\%$ of the $\tfrac18$ budget (\cref{eq:prefix}),
leaving little capacity after pruning.
\Cref{fig:qualitative} shows how \method retains queried object representations through successive reductions at $B=\tfrac1{16}$.

\noindent\textbf{Position robustness and joint evidence.}
Two controlled probes test whether the equal per-image quota preserves evidence accessibility
(\cref{tab:probes}). \textbf{(1) Position.} When the relevant MMStar~\cite{NEURIPS2024_2f8ee6a3}
image moves among $N-1$ distractors, \method has the highest mean and worst-position accuracy,
$3.6$ and $4.8$ points above the best baseline in mean accuracy at $N=4$ and $8$.
\textbf{(2) Joint use.} On COCO image pairs that need a fact from each image, \method reaches
$86.6\%$ and $77.8\%$ joint accuracy at $N=4$ and $8$, above both the best baselines and the
uncompressed model. When either image is replaced so that the answer changes, it answers both
versions correctly about twice as often as the uncompressed model at $N=8$. Replacing equal quotas with global top-$k$ or a minimum quota plus attention-based allocation reduces most metrics, supporting equal per-image allocation. Together, these controlled probes provide evidence that equal quotas improve access to relevant visual information across image positions and support its joint use in cross-image reasoning. Protocols and more results are in supplementary.

\begin{table}[t]
\centering
\caption{\textbf{Position robustness and joint evidence.} Mean and Worst are the average and lowest accuracy over relevant image positions. Joint is the accuracy on questions that need one fact from each of two images. CF-A and CF-B are the share of pairs answered correctly both before and after image A or B is replaced so that the answer changes (N = 8).}
\label{tab:probes}
\renewcommand{\arraystretch}{0.9}
\resizebox{\linewidth}{!}{%
\begin{tabular}{lcccc|cccc}
\toprule
& \multicolumn{4}{c|}{\cellcolor{bbllava}\textbf{Position}} & \multicolumn{4}{c}{\cellcolor{bbllava}\textbf{Joint evidence}} \\
& \multicolumn{2}{c}{\cellcolor{bbqwenv3}{$N=4$}} & \multicolumn{2}{c|}{\cellcolor{bbqwenv3}{$N=8$}} & \cellcolor{bbqwenv3}{$N=4$} & \multicolumn{3}{c}{\cellcolor{bbqwenv3}{$N=8$}} \\
\cmidrule(lr){2-3}\cmidrule(lr){4-5}\cmidrule(lr){6-6}\cmidrule(l){7-9}
Method & Mean $\uparrow$ & Worst $\uparrow$ & Mean $\uparrow$ & Worst $\uparrow$ & Joint $\uparrow$ & Joint $\uparrow$ & CF-A $\uparrow$ & CF-B $\uparrow$ \\
\midrule
\textit{Full}         & 52.9 & 51.2 & 48.1 & 45.2 & 77.1 & 63.9 & 23.6 & 23.2 \\
\midrule
VisionZip~\cite{Yang_2025_CVPR}             & 43.1 & 41.2 & 40.1 & 37.8 & 67.2 & 60.7 & 20.5 & 19.8 \\
FastV~\cite{Chen_2024}                      & 30.5 & 29.0 & 30.0 & 27.8 & 50.8 & 44.0 & 2.2 & 2.3 \\
PyramidDrop~\cite{Xing_2025_CVPR}           & 28.8 & 27.0 & 28.6 & 26.8 & 49.9 & 42.0 & 2.7 & 3.2 \\
SparseVLM~\cite{pmlr-v267-zhang25s}         & 30.4 & 28.8 & 29.8 & 27.4 & 51.9 & 43.7 & 4.2 & 4.9 \\
AVTP~\cite{zhang2026multiimagevisualtokenpruning} & 33.5 & 32.0 & 33.5 & 32.0 & 58.0 & 54.6 & 0.7 & 1.0 \\
CLSE~\cite{chen2026spectralevolutionguidedtokenpruning}& 38.3 & 33.0 & 34.8 & 30.0 & 51.7 & 43.9 & 6.2 & 6.5 \\
S$^2$Prune~\cite{jia2026s2prunespatiallystructuredvisual} & 44.0 & \underline{43.0} & 40.4 & \underline{38.2} & 64.3 & 54.1 & 13.6 & 12.8 \\
ToDRE~\cite{li2026todre}                    & 42.0 & 40.4 & 39.7 & 37.6 & 68.2 & 59.3 & 20.9 & 19.4 \\
\midrule
\method w/ global top-$k$      & \underline{45.8} & \underline{43.0} & \underline{43.3} & 37.4 & \underline{85.0} & \underline{74.6} & \underline{42.0} & 37.1 \\
\method w/ floor + attention   & 45.1 & 41.4 & 43.2 & 36.6 & 84.0 & \underline{74.6} & 41.5 & \underline{37.4} \\
\rowcolor{ourshade}
\textbf{\method} & \textbf{47.6} & \textbf{47.0} & \textbf{45.2} & \textbf{42.2} & \textbf{86.6} & \textbf{77.8} & \textbf{47.3} & \textbf{42.5} \\
\bottomrule
\end{tabular}%
}
\vspace{-3mm}
\end{table}

\noindent\textbf{Larger image sets.}
On MuirBench prompts with $N\geq5$ (supplementary D2), \method has the highest four-backbone mean at $B=\tfrac1{16}$
($51.87$ vs.\ $50.05$ for ToDRE).

\noindent\textbf{Transfer to video understanding.}
Video presents a distinct compression regime from static multi-image reasoning: adjacent frames are strongly correlated and contain substantial visual redundancy. For Video-MME~\cite{fu2025video} and EgoSchema~\cite{mangalam2023egoschema}, we use each backbone's native $32$-frame video pipeline and treat each frame---or each temporal patch for the Qwen models---as one observation for compression. \method requires no video-specific modules, retraining, or hyperparameter changes; it applies the same spatial pooling, per-image allocation, and progressive select-and-merge schedule. Across four backbones, \method achieves the highest mean among the evaluated compression methods at both $B=\tfrac1{16}$ and $\tfrac18$, reaching $59.76$ and $61.21$ and exceeding ToDRE by $0.79$ and $0.51$ points, respectively (\cref{tab:video}). These results demonstrate direct transfer from heterogeneous image sets to correlated frame sequences.

\noindent\textbf{Transfer to embodied applications.}
Embodied tasks expose visual context through heterogeneous observation structures, including individual observations, temporal sequences, and multi-camera views. \method processes these inputs through the same interface and configuration, requiring no task-specific adaptation beyond representing each observation as an image. Across five embodied benchmarks and four backbones, \method achieves the highest mean at both budgets, reaching $39.00$ at $B=\tfrac1{16}$ and $40.29$ at $B=\tfrac18$ (\cref{tab:robotics}). These scores exceed the strongest corresponding baselines by $0.45$ and $0.79$ points.

Consistent performance across complementary image sets, correlated video frames, and embodied visual observations demonstrates that \method generalizes across diverse observation structures without task-specific modifications to its compression pipeline. \method thus provides a unified framework for multi-observation compression.

\subsection{Single-image results}
\label{sec:results-single}

For a single image, the per-image quota reduces to global top-$k$ and the remaining operations are unchanged.
Across four backbones and four benchmarks, \method averages $62.96$ and $69.59$ at $B=\tfrac1{16}$ and $\tfrac18$
($77\%$ and $85\%$ of the uncompressed), the highest four-backbone average at both budgets and $2.91$ and $1.07$ points above the strongest baseline, VisionZip (\cref{tab:single-mean}).

\subsection{Efficiency}
\label{sec:efficiency}
\Cref{tab:efficiency} reports realized cost on LLaVA-OneVision-7B at $B=\tfrac18$ (five-benchmark average on one H100).
\method cuts prefill computation from $26.46$ to $4.02$\,TFLOPs (\textbf{$6.6\times$}) and KV-cache storage from $107.4$ to $16.7$\,MB (\textbf{$6.4\times$}), both the lowest of all methods.
At the same budget, all methods have FLOPs and storage within $3\%$ of SIEVE's, yet \method is $9.37$ points more accurate than the best baseline.
Prefill latency falls from $100.4$ to $53.5$\,ms ($1.9\times$), within $6.9$\,ms of the fastest method, which is $20.91$ points less accurate.
\begin{table}[t]
  \centering
  \caption{\textbf{Ablation study of \method.} Each variant modifies one component of the full method at $B=\tfrac1{16}$. Scores are averaged across LLaVA-OneVision-7B, Qwen2.5-VL-7B, Qwen3-VL-8B, and InternVL3-8B. \textbf{Avg} is the mean across five benchmarks; $\Delta$ denotes the change from full \method.}
  \label{tab:ablation}
  \setlength{\tabcolsep}{2.1pt}
  \renewcommand{\arraystretch}{0.92}
  \resizebox{\columnwidth}{!}{
  \begin{tabular}{@{}lccccc>{\columncolor{avgshade}}rr@{}}
    \toprule
    & \cellcolor{sectshade}\textbf{Mantis} & \cellcolor{sectshade}\textbf{NLVR2} & \cellcolor{sectshade}\textbf{Muir} & \cellcolor{sectshade}\textbf{BLINK} & \cellcolor{sectshade}\textbf{Q-Bench2} & \textbf{Avg} & $\boldsymbol{\Delta}$ \\
    \midrule
    \rowcolor{ourshade} \textbf{\method} & \textbf{65.90} & \textbf{78.29} & \textbf{49.23} & \textbf{47.20} & \textbf{69.70} & \textbf{62.06} & --- \\
    \midrule
    \rowcolor{supershade} \multicolumn{8}{@{}l}{\textbf{RQ1: Cross-image coverage and prompt-guided selection}} \\
    global top-$k$ (no per-image quota) & 62.79 & 76.94 & 46.15 & 45.37 & 66.98 & 59.65 & \dbad{$-2.41$} \\
    random selection scores & 61.75 & 76.70 & 46.07 & 44.27 & 65.15 & 58.79 & \dbad{$-3.27$} \\
    \midrule
    \rowcolor{supershade} \multicolumn{8}{@{}l}{\textbf{RQ2: Token aggregation and image-identity preservation}} \\
    drop unselected tokens & 64.52 & 77.50 & 46.25 & 45.67 & 67.13 & 60.21 & \dbad{$-1.85$} \\
    allow cross-image merge targets & 64.52 & 77.71 & 46.10 & 45.52 & 67.00 & 60.17 & \dbad{$-1.89$} \\
    spatially nearest same-image target & 64.40 & 77.04 & 45.84 & 45.70 & 66.13 & 59.82 & \dbad{$-2.24$} \\
    \midrule
    \rowcolor{supershade} \multicolumn{8}{@{}l}{\textbf{RQ3: Initial pooling and progressive reduction}} \\
    no pooling (selection at layer $0$) & 54.96 & 64.53 & 40.02 & 44.12 & 63.90 & 53.51 & \dbad{$-8.55$} \\
    pooling only & 62.56 & 75.18 & 44.11 & 44.58 & 64.03 & 58.09 & \dbad{$-3.97$} \\
    pooling + FastV & 61.41 & 73.71 & 44.12 & 46.27 & 66.35 & 58.37 & \dbad{$-3.69$} \\
    pooling + PyramidDrop & 63.25 & 72.61 & 43.57 & 47.07 & 68.03 & 58.90 & \dbad{$-3.16$} \\
    \midrule
    \rowcolor{supershade} \multicolumn{8}{@{}l}{\textbf{RQ4: Depth-wise placement of reduction stages}} \\
    first stage at layer $0$ & 64.29 & 77.55 & 46.60 & 45.70 & 66.48 & 60.12 & \dbad{$-1.94$} \\
    first stage at layer $8$ & 61.64 & 72.76 & 44.27 & 44.56 & 66.68 & 57.98 & \dbad{$-4.08$} \\
    later stages evenly spaced & 64.40 & 78.22 & 44.91 & 45.74 & 67.45 & 60.14 & \dbad{$-1.92$} \\
    later stages in the first half & 62.33 & 74.81 & 42.01 & 44.55 & 67.00 & 58.14 & \dbad{$-3.92$} \\
    \bottomrule
  \end{tabular}%
  }
\end{table}
\begin{table}[!t]
  \centering
  \caption{\textbf{Reduction schedule and pooling.} Mean multi-image accuracy; ---: infeasible.}
  \label{tab:schedule-pool}
  \setlength{\tabcolsep}{4pt}
  \renewcommand{\arraystretch}{0.95}
  \resizebox{0.8\columnwidth}{!}{%
  \begin{tabular}{@{}lcccc@{}}
    \toprule
    & \multicolumn{2}{c}{\cellcolor{sectshade}\textbf{Keep 6.25\%}} & \multicolumn{2}{c}{\cellcolor{sectshade}\textbf{Keep 12.5\%}} \\
    \cmidrule(lr){2-3} \cmidrule(l){4-5}
    & \cellcolor{bbllava}\textbf{OneVision} & \cellcolor{bbqwenv2}\textbf{Qwen2.5} & \cellcolor{bbllava}\textbf{OneVision} & \cellcolor{bbqwenv2}\textbf{Qwen2.5} \\
    \midrule
    \rowcolor{ourshade} \textbf{\method} & \textbf{62.19} & \textbf{63.67} & \textbf{65.13} & \textbf{66.21} \\
    Recursive, $\ell = 1$ & 58.05 & 60.75 & 63.74 & 63.76 \\
    Recursive, $\ell = 2$ & 57.89 & 60.59 & 64.03 & 63.09 \\
    Recursive, $\ell = 3$ & 55.81 & 58.08 & 62.01 & 61.90 \\
    \midrule
    No pooling, $\ell_1 = 0$ & 46.51 & 56.14 & 51.75 & 63.77 \\
    No pooling, $\ell_1 = 2$ & --- & --- & 46.64 & 55.67 \\
    \bottomrule
  \end{tabular}%
  }
  \vspace{-3mm}
\end{table}

\subsection{Ablations}
\label{sec:ablations}
We investigate four questions through ablation studies: the role of per-image allocation and prompt-guided selection, the benefit of same-image token aggregation, the contributions of initial pooling and progressive reduction, and the effect of reduction-stage placement (\cref{tab:ablation}). Results are averaged across four backbones and five multi-image benchmarks at $B=\tfrac1{16}$. Additional ablations and per-backbone results are provided in the supplementary material.

\noindent\textbf{RQ1: Cross-image coverage and prompt-guided selection.}
Replacing equal per-image quotas with global top-$k$ selection reduces the mean score by $2.41$ points, with losses of $2.08$, $2.61$, $1.44$, and $3.54$ points on LLaVA-OneVision, Qwen2.5-VL, Qwen3-VL, and InternVL3, respectively. Randomizing selection within the same quotas reduces the mean by $3.27$ points. These results demonstrate that both balanced allocation across images and prompt-guided selection within each image contribute to effective compression.

\noindent\textbf{RQ2: Token aggregation and image-identity preservation.}
Discarding unselected tokens reduces the mean score by $1.85$ points, demonstrating the benefit of aggregation. Allowing cross-image merging reduces the mean by $1.89$ points, with losses of $1.09$, $2.06$, $0.96$, and $3.48$ points on LLaVA-OneVision, Qwen2.5-VL, Qwen3-VL, and InternVL3, respectively. Replacing feature-similarity matching with spatially nearest targets within each image reduces the mean by $2.24$ points. These results support aggregating unselected tokens into semantically similar representatives from the same image.

\noindent\textbf{RQ3: Initial pooling and progressive reduction.}
Removing pooling and performing the first selection at layer~$0$ reduces the mean score by $8.55$ points, with losses of $15.68$, $7.53$, $5.02$, and $6.00$ points on LLaVA-OneVision, Qwen2.5-VL, Qwen3-VL, and InternVL3, respectively. Pooling alone achieves $58.09$, while subsequent progressive selection and merging improves the mean to $62.06$, yielding gains of $3.70$, $3.90$, $2.87$, and $5.41$ points on the four backbones. Applying FastV or PyramidDrop after pooling achieves $58.37$ and $58.90$, respectively, remaining $3.69$ and $3.16$ points below \method. These results demonstrate the complementary benefits of initial spatial pooling and progressive selection and aggregation. The benefit persists at a $12.5\%$ budget: removing pooling reduces accuracy by $13.38$ points on LLaVA-OneVision and $2.44$ points on Qwen2.5-VL (\cref{tab:schedule-pool}).

\noindent\textbf{RQ4: Depth-wise placement of reduction stages.}
Moving the first reduction from layer~$2$ to layer~$0$ reduces the mean score by $1.94$ points, while delaying it to layer~$8$ incurs a $4.08$-point loss. For subsequent reductions, evenly spacing stages across decoder depth reduces the mean by $1.92$ points, and placing them in the first half reduces it by $3.92$ points. These results support an early reduction to control processing cost, followed by later stages that select and aggregate tokens after deeper multimodal processing.

\noindent\textbf{Progressive versus recursive reduction.}
We additionally evaluate recursive reduction on two backbones, LLaVA-OneVision-7B and Qwen2.5-VL-7B (\cref{tab:schedule-pool}). The recursive variant selects and merges at a fixed layer, then restarts inference from the first layer; its repeated scoring passes count toward the budget. Relative to the strongest recursive configuration for each backbone and budget, \method improves the two-backbone mean by $3.53$ points at $6.25\%$ and $1.78$ points at $12.5\%$. The corresponding per-backbone gains are $4.14$ and $2.92$ points at $6.25\%$, and $1.10$ and $2.45$ points at $12.5\%$, respectively. These results highlight the benefit of deferring part of the reduction to later decoder layers.
\vspace{-1mm}
\section{Conclusion}
\label{sec:conclusion}
\vspace{1mm}
We present \method, a training-free, single-pass framework for multi-image token compression. It pools images before decoding, then selects and merges tokens at selected layers under equal per-image quotas and a depth-averaged budget. At a $6.25\%$ budget, it achieves the highest average across four backbones on five multi-image, four single-image, two video, and five embodied application benchmarks. These results highlight the importance of jointly optimizing compression timing, prompt-guided selection under equal per-image quotas, and same-image token merging.

\section*{LLM Usage Statement}
Claude was used to assist with writing and reviewing code, while GPT was used to assist with manuscript writing and language polishing. The authors reviewed and verified all generated content and take full responsibility for the code and manuscript.

\clearpage
\maketitlesupplementary
\appendix

\setcounter{topnumber}{2}
\section{Motivation}
\label{sec:appendix-motivation}

\subsection{Prompt attention across decoder depth}
\label{sec:appendix-globalrank}

\noindent\textbf{Attention to the queried object} (Fig.~1(a), upper).
We use COCO val2017 \cite{Lin_2014} images whose largest non-crowd instance covers $3$--$60\%$ of the
image: $400$ images for LLaVA-OneVision \cite{li2024llavaonevisioneasyvisualtask} and Qwen3-VL \cite{bai2025qwen3vltechnicalreport}, and $249$ for InternVL3 \cite{zhu2025internvl3exploringadvancedtraining}. Each image
is the only image in the prompt, and the question is ``What is the $\langle$category$\rangle$
doing in the image?''. At every layer, we measure the share of the final prompt token's
visual-token attention that falls inside the instance mask. As a control, we also measure the
share inside a centered box of the same area, because photographs and attention are both biased
toward the image center.

\noindent\textbf{Global top-$10\%$ ranking} (Fig.~1(a), lower).
We test whether a global attention ranking over all images can guide early
token reduction. We use 500 four-image sets from the COCO joint-evidence
probe (\cref{sec:pos_joint}), with the relevant images at positions 1 and 3.
Each set is evaluated with up to seven questions: the three probe questions
(one about each relevant image and the joint question) and four
image-specific questions of the form ``Is there a $\langle$object$\rangle$
in image $p$?'', yielding 3{,}485 questions per backbone (images without an
annotated object are excluded from these image-specific queries). We rank all pooled
visual tokens by the final prompt token's attention at five decoder depths
and at the last layer, and select the global top-$10\%$.
\Cref{tab:globalrank} reports the mean pairwise Jaccard overlap among the questions asked
about the same image set, which target different images (random: $5\%$), and the share of
selected tokens in the queried image (uniform: $25\%$). Bootstrap 95\% confidence intervals (10,000 image-set resamples) remain within $\pm0.7$ points. In the first third of the decoder, rankings are largely query-agnostic, with $81$--$86\%$ overlap and near-uniform allocation to the queried image. Only in the second half does the queried image receive most of the selected tokens, with per-backbone peaks of $51$--$72\%$. At the final layer, LLaVA-OneVision and InternVL3 retain this preference ($63\%$ and $45\%$), whereas Qwen3-VL reverts toward uniform allocation ($26\%$, $73\%$ overlap). At layer~2 the selection is also position-biased: $49\%$ falls on the last image for LLaVA-OneVision and $63\%$ on the first for Qwen3-VL (uniform: $25\%$).

\begin{table}[t]
  \centering
  \caption{\textbf{Global top-10\% by relative decoder depth.} Overlap: mean pairwise Jaccard among the
  questions asked about the same image set (random: $5\%$). Queried-image share: fraction of selected tokens within the queried image (uniform: $25\%$). 500 image sets per backbone. Relative depth is $\ell/(L-1)$; the column
  headers are the values for the 28-layer backbones (layers 2, 8, 14, 19, 23,
  and the last). Qwen3-VL (layers 2, 10, 18, 24, 30, and the last of 36) has
  relative depths $0.06, 0.29, 0.51, 0.69, 0.86, 1.00$. Bold marks each backbone's peak queried-image share.}
  \label{tab:globalrank}
  \scriptsize
  \setlength{\tabcolsep}{3pt}
  \resizebox{\columnwidth}{!}{%
  \begin{tabular}{llcccccc}
    \toprule
    & & \multicolumn{6}{c}{\cellcolor{sectshade}\textbf{relative depth}} \\
    \cmidrule(lr){3-8}
    \textbf{Backbone} & & \cellcolor{avgshade}\textbf{0.07} & \cellcolor{avgshade}\textbf{0.30} & \cellcolor{avgshade}\textbf{0.52} & \cellcolor{avgshade}\textbf{0.70} & \cellcolor{avgshade}\textbf{0.85} & \cellcolor{avgshade}\textbf{1.00} \\
    \midrule
    \cellcolor{bbllava}\textbf{LLaVA-OneVision-7B} & Overlap & 82.5 & 81.0 & 44.8 & 25.0 & 23.0 & 25.2 \\
    \cellcolor{bbllava} & \makecell{Queried-image\\share} & 24.9 & 25.4 & 44.3 & 69.9 & \textbf{70.0} & 62.7 \\
    \midrule
    \cellcolor{bbqwenv3}\textbf{Qwen3-VL-8B} & Overlap & 85.3 & 85.7 & 81.0 & 22.7 & 53.1 & 73.2 \\
    \cellcolor{bbqwenv3} & \makecell{Queried-image\\share} & 25.1 & 25.0 & 28.6 & \textbf{71.9} & 38.8 & 26.1 \\
    \midrule
    \cellcolor{bbintern}\textbf{InternVL3-8B} & Overlap & 84.4 & 82.6 & 67.0 & 31.7 & 35.7 & 35.7 \\
    \cellcolor{bbintern} & \makecell{Queried-image\\share} & 24.9 & 25.2 & 29.8 & \textbf{51.5} & 48.8 & 45.0 \\
    \bottomrule
  \end{tabular}%
  }
\end{table}

\subsection{Component contributions}
\label{sec:appendix-ladder}

In Fig.~1(b), adding a single merge stage to selection with per-image quotas
improves mean multi-image accuracy by $0.04$ points, and repeating selection
and merging across layers adds a further $2.27$ points (LLaVA-OneVision-7B,
$B=\tfrac1{16}$).

\section{Method details}
\label{sec:appendix-method}

\subsection{Algorithm}
\label{sec:appendix-algorithm}

\Cref{alg:sieve} summarizes the \method prefill procedure.

\begin{algorithm}[!tp]
\caption{SIEVE: multi-image token compression}
\label{alg:sieve}
\small
\begin{algorithmic}[1]
\Require Visual grids $\{X_i\}_{i=1}^{N}$, prompt $\mathbf{p}$, decoder $\{f_\ell\}_{\ell=0}^{L-1}$, target budget $B$
\Ensure Compressed prefill states and layer-wise KV caches
\State Set $\alpha=3$, $\beta=0.75$, $\gamma=\tfrac12$, $\ell_1=2$, $q_{\min}=4$
\Statex \textbf{1. Spatial pooling before the decoder}
\State $P_i\gets |X_i|$; $n\gets\sum_i P_i$
\State $m\gets\lfloor\min(\alpha B,\beta)n\rfloor$
\For{$i=1,\ldots,N$}
    \State $m_i\gets\operatorname{round}(mP_i/n)$
    \State Resample $X_i$ to approximately $m_i$ tokens using the backbone-specific operator
\EndFor
\State Interleave pooled grids with $\mathbf{p}$ to form $H$; initialize owner map $\omega$ and position indices
\Statex \textbf{2. Image-adaptive, budget-aware scheduling}
\State $S\gets\min(4,\max(2,N))$
\State Compute $\ell_2,\ldots,\ell_S$ using Eq.~(3)
\State Set $\ell_{S+1}\gets L-1$ and $\delta_j\gets\ell_{j+1}-\ell_j$
\State $k_1\gets\displaystyle\frac{BnL-m(\ell_1+1)}{\sum_{j=1}^{S}\gamma^{j-1}\delta_j}$
\State $k_j\gets\operatorname{clip}\big(\operatorname{round}(k_1\gamma^{j-1}),1,m\big)$ for each stage $j$
\For{$\ell=0,\ldots,L-1$}
    \State Run $f_\ell$ on $H$, retaining its outputs and KV cache
    \If{$\ell=\ell_j$ for some stage $j$}
        \Statex \hspace{\algorithmicindent}\textbf{3. Prompt-attention scoring}
        \State Compute $s_t$ from the final prompt token's attention in $f_\ell$, normalized over visual tokens per head (Eq.~(4))
        \Statex \hspace{\algorithmicindent}\textbf{4. Per-image coverage}
        \State Set $\mathcal V_i\gets\{t:\omega(t)=i\}$
        \State Allocate $k_j$ slots uniformly across images with floor $q_{\min}$ and capacity $|\mathcal V_i|$; distribute the remainder (Sec.~3.3)
        \State $\mathcal K_i\gets\operatorname{TopK}(\{s_t:t\in\mathcal V_i\},q_i)$; $\mathcal K\gets\bigcup_i\mathcal K_i$
        \Statex \hspace{\algorithmicindent}\textbf{5. Within-image evidence merging}
        \State Using layer outputs, assign each unselected $t$ to $\pi(t)\gets\argmax_{u\in\mathcal K_{\omega(t)}}\cos(\mathbf h_t,\mathbf h_u)$
        \State Update all representatives simultaneously by score-weighted averaging (Eq.~(5))
        \Statex \hspace{\algorithmicindent}\textbf{6. Sequence and position update}
        \State Keep all text tokens and merged representatives in sequence order; retain their owner and position indices
        \State Pass the reduced sequence to the next layer; retain the caches already computed
    \Else
        \State Pass the layer outputs to the next layer
    \EndIf
\EndFor
\State \Return Final prefill states and layer-wise KV caches
\end{algorithmic}
\end{algorithm}

\subsection{Backbone-specific pooling}
\label{sec:appendix-pooling}

Image $i$ receives $m_i = \operatorname{round}(m P_i / n)$ tokens from the pooling target
(Sec.~3.2). Coordinates $c_t\in\mathbb{R}^2$ denote rows and columns on the
unpooled grid; pooled coordinates $\tilde c$ denote cell centers in the same frame.

\noindent\textbf{LLaVA-OneVision, multi-image.}
Each image has a $27\times27$ grid ($G=27$) and receives
$m_i=\operatorname{round}(m/N)$ tokens. Adaptive average pooling gives
$\tilde X_i=\avgpool_{\tilde G\times\tilde G}(X_i)$ with
$\tilde G=\operatorname{round}(\sqrt{m_i})$. Cell $(a,b)$ has center
$\tilde c=\big((a+\tfrac12)G/\tilde G-\tfrac12,\;(b+\tfrac12)G/\tilde G-\tfrac12\big)$.
Because $\tilde G$ is an integer, the realized total
$\tilde m=\sum_i\tilde G^2$ may differ slightly from $m$
(\cref{sec:appendix-stages}).

\noindent\textbf{LLaVA-OneVision, single image.}
AnyRes produces a variable-size token grid from several image crops. We
bin its tokens by (row, column) into the smallest $\tilde G\times\tilde G$
grid with $\tilde G\geq\operatorname{round}(\sqrt{m_i})$ and at least $m_i$
non-empty cells, average the features within each non-empty cell, and retain
the row-separator (newline) tokens that LLaVA-OneVision appends to each row.

\noindent\textbf{Qwen2.5-VL.} \cite{bai2025qwen25vltechnicalreport}
The $G_h\times G_w$ grid is pooled to
$\operatorname{round}(\sqrt{m_iG_h/G_w})\times\operatorname{round}(\sqrt{m_iG_w/G_h})$ cells,
preserving its aspect ratio up to rounding and using the cell centers defined above.

\noindent\textbf{Qwen3-VL.}
Qwen3-VL grids are small, so this rounding can lose up to about $20\%$ of $m_i$. We instead
choose the $\tilde G_h\times\tilde G_w$ grid, with $\tilde G_h\le G_h$ and $\tilde G_w\le G_w$,
whose cell count is closest to $m_i$, subject to an aspect ratio within $1.5\times$ of $G_h/G_w$.

\noindent\textbf{InternVL3.}
Each image is split into up to $\max(1,\min(6,\lfloor 64/N\rfloor))$ tiles
of $448\times448$ pixels, plus one $16\times16$-token thumbnail. We pool the
tile mosaic and the thumbnail separately, allocating each a share of the pooled
budget proportional to its token count, then concatenate the pooled representations, with the mosaic
preceding the thumbnail.

\subsection{Position indices after reduction}
\label{sec:appendix-positions}

Pooled tokens follow raster order, with text, row separators, and image markers preserving
their original order (Sec.~3.3). Cell centers are stored separately from
position indices for spatial bookkeeping, including AnyRes binning, per-image quotas,
neighbor search, and merge-target selection.

\noindent\textbf{LLaVA-OneVision and InternVL3.}
Both use one-dimensional rotary embeddings with consecutive positions $0,\dots,T-1$
on the pooled sequence. InternVL3 thumbnail coordinates are rescaled to the mosaic frame
for spatial bookkeeping only.

\noindent\textbf{Qwen2.5-VL and Qwen3-VL.}
After pooling, we reconstruct the three M-RoPE components: text runs use
consecutive positions in all three components, and image blocks use $(0,r,c)$, with the running
offset advancing by $\max(r,c)+1$. Here, $r$ and $c$ are dense ranks of the distinct
row and column coordinates within each image. Rank compression supports non-rectangular
token sets, whereas the official index builders assume a dense $G_h\times G_w$ grid.

\noindent\textbf{Decoder stages.}
Surviving tokens retain their indices, introducing gaps as tokens are
removed (\eg, $1,2,3,4\rightarrow 2,4$). For M-RoPE, every \method stage
uses positions from the table constructed before layer~0. Each representative retains its own index and coordinates after
aggregation.

\noindent\textbf{Position ablations.}
\Cref{tab:ablation-extra} tests two alternatives. In ``average the positions of merged
tokens'', each surviving token's position is the average of the positions
of all tokens it absorbed, weighted by the content weights of Eq.~(5).
Fractional rotary positions are stored separately from the integer cache
indices, which only the LLaVA-OneVision implementation supports. In ``keep pre-pooling
indices'', pooled tokens keep the indices of the unpooled grid instead of being renumbered.

\subsection{Stage count and schedule shape}
\label{sec:appendix-stages}

\noindent\textbf{Worked example.}
For LLaVA-OneVision with $N=4$, $n=2916$, and $B=\tfrac18$, Eq.~(6) gives

\begin{center}
\begin{tabular}{lcc}
\toprule
Stage & Token reduction & Per image \\
\midrule
Pooling (before $f_0$) & $2916\!\to\!1156$ & 289 \\
After $f_2$     & $1156\!\to\!372$  & 93 \\
After $f_{18}$  & $372\!\to\!186$   & 46--47 \\
After $f_{21}$  & $186\!\to\!93$    & 23--24 \\
After $f_{24}$  & $93\!\to\!46$     & 11--12 \\
\bottomrule
\end{tabular}
\end{center}

The depth-averaged count is $371.25$, compared with the target $Bn=364.5$.
The pooling target $m=1093$ gives $m_i=273$, rounded to a $17\times17$ grid per image
($\tilde m=1156$). The 63 additional tokens across three prefix layers contribute
$6.75$ depth-averaged tokens, or $1.9\%$ of the budget. At $B=\tfrac1{16}$,
$m=546$, $m_i=136$, and $\tilde G=12$ yield $\tilde m=576$, adding $1.8\%$
before subsequent retained counts are rounded.
\Cref{tab:ablation-extra} compares fixed stage counts with the adaptive schedule $S(N)$.

\noindent\textbf{Schedule variants.}
The two placement variants of Table~6 either space all stages evenly after $\ell_1$ or confine them to
the first half of the remaining depth. \Cref{tab:ablation-extra} also varies the decay $\gamma$, the pooling headroom
$\alpha$, the first-stage layer $\ell_1$, and the schedule shape one at a time. The constant-ratio variant keeps
$k_j=m\,r^{j}$ tokens after stage $j$, as in PyramidDrop, instead of the
geometric schedule of Eq.~(6); $r$ is solved numerically so that
the depth-averaged count meets the budget.
The cap $\beta$ is not varied: it binds only when $\alpha B>0.75$, so it is
inactive at every budget we report.

\section{Evaluation details}
\label{sec:appendix-evaluation}

\subsection{Backbones and baselines}
\label{sec:appendix-backbones}

We use the official implementations and configurations of the evaluated methods
when available, adapting only the backbone-dependent quantities required by differences in
decoder depth, visual tokenization, or vision-encoder architecture. All \method results use the
shared hyperparameters in Sec.~4.1; across backbones, only the decoder depth $L$ changes the
stage placement (Eq.~(3)). For Qwen3-VL-8B ($L=36$), the two-, three-, and four-stage schedules place
reductions after layers $\{2,27\}$, $\{2,24,30\}$, and $\{2,23,27,31\}$, respectively.

We likewise preserve the published depth-based placement rules of the baselines. On Qwen3-VL,
PyramidDrop \cite{Xing_2025_CVPR} and SparseVLM \cite{pmlr-v267-zhang25s} operate at layers $\{2,9,18,27\}$ and $\{2,7,17\}$, respectively;
on InternVL3, both use their configurations for 28-layer decoders. VisionZip \cite{Yang_2025_CVPR} retains its original
dominant--contextual token construction and adapts only the attention source to each vision
encoder. The Qwen backbones use attention received in the final ViT block, aggregated over each
$2\times2$ merged patch group; InternVL3 uses class-token attention, aggregated over the patch
groups combined by pixel shuffle; and LLaVA-OneVision uses patch-level received attention because
its SigLIP~\cite{siglip} encoder has no class token. Contextual tokens are formed using attention-key
similarity, with the published dominant-to-contextual allocations for the corresponding encoder.

\subsection{Feasibility constraints on token budgets}
\label{sec:appendix-infeasible}

A method whose first reduction follows layer $\ell_1$ spends
$(\ell_1+1)/(BL)$ of the target budget before pruning (Eq.~(2)). If
$BL<\ell_1+1$, no later pruning can meet the target.
\begin{itemize}
  \item \textbf{SparseVLM} ($\ell_1=2$) cannot meet $6.25\%$ on any
        backbone: its solver returns no schedule, so it is omitted from all $6.25\%$ rows.
  \item \textbf{FastV \cite{Chen_2024} and PyramidDrop} ($\ell_1=2$) fail the same
        criterion. Their retained token counts are clamped to
        one token, so they remain executable but exceed the target budget ($\ddagger$ in the tables).
  \item \textbf{CLSE} \cite{chen2026spectralevolutionguidedtokenpruning} ($\ell_1=1$) needs $B\geq 2/L$, \ie, $7.14\%$ on the
        three 28-layer backbones and $5.56\%$ on Qwen3-VL. It therefore
        runs at $6.25\%$ only on Qwen3-VL.
  \item \textbf{S$^2$Prune} \cite{jia2026s2prunespatiallystructuredvisual}: seven Qwen2.5-VL multi-image prompts (three
        BLINK, four MuirBench) are infeasible at $6.25\%$. Excluding them
        changes each of those two cells by less than $0.1$ points. On the
        same backbone and budget, $87$ MMStar and $51$ ScienceQA-IMG prompts
        are also infeasible, and those two single-image cells are scored on
        the feasible prompts.
\end{itemize}
In the main paper's Tables~2 and~3 and in \cref{tab:single,tab:n5,tab:video-full,tab:robotics-full}, a dash
denotes an infeasible or unevaluated configuration.

\subsection{Benchmarks and metrics}
\label{sec:appendix-metrics}
\label{sec:appendix-tables}

\Cref{tab:metrics} summarizes the evaluated splits, number of questions (\#Q), number of input images or frames per question (\#Img), and evaluation metrics. All splits are evaluated in full, except prompts
excluded as infeasible; Q-Bench2 uses its publicly labeled development split. We use greedy decoding and the same answer parser for all methods within each benchmark. \textit{Full} denotes the uncompressed reference.

\noindent\textbf{Multi-image benchmarks.}
We evaluate multiple-choice questions using accuracy, with the first standalone option letter in the response as the prediction. Mantis-Eval also includes short-answer questions, which we score using normalized containment matching: a prediction is correct if either the response or reference contains the other.

\noindent\textbf{Single-image benchmarks.}
ScienceQA-IMG and MMStar use the same option-letter accuracy. TextVQA follows the standard VQA metric after official answer normalization, averaging $\min(\#\text{matches}/3,1)$ over the ten leave-one-out annotator subsets. For POPE, we report mean accuracy across its random, popular, and adversarial splits, classifying a response as \emph{yes} if it begins with \emph{yes}.

\noindent\textbf{Video benchmarks.}
We follow the backbone-specific video evaluation protocols implemented in \texttt{lmms-eval}. Each clip is represented by $32$ frames selected using the corresponding backbone's native frame-indexing rule and processed through its standard video pipeline. LLaVA-OneVision pools each frame to $196$ visual tokens and appends one newline token per clip. Qwen2.5-VL and Qwen3-VL combine adjacent frame pairs into temporal patches, with Qwen3-VL additionally providing textual timestamps. InternVL3 encodes each frame as a single $448\times448$ tile. We use the \texttt{lmms-eval} prompts and answer parsing for Video-MME and the labeled subset of EgoSchema, with greedy decoding and a maximum of $16$ generated tokens. Compression budgets are defined relative to each backbone's native visual-token count. For compression, all methods treat each frame---or each temporal patch for the Qwen models---as one image.

\noindent\textbf{Embodied application benchmarks.}
ERQA and MV-RoboBench are evaluated using option-letter accuracy. RoboSpatial is evaluated on its Home split (RoboSpatial-Home) and reports accuracy macro-averaged across question types based on the leading yes/no response, excluding pointing (\emph{context}) questions that require point-in-mask evaluation. NuScenes-QA uses normalized exact or whole-token matching, macro-averaged across five question types. For OpenEQA, we uniformly sample $16$ frames per episode and follow the official LLM-Match evaluation protocol, using Qwen2.5-32B-Instruct as the judge.

\begin{table}[t]
  \centering
  \caption{\textbf{Per-benchmark statistics: split, question count, image count, and metrics.} \#Q is the number of questions, and \#Img is the number of images per prompt, reported as minimum--maximum (mean) when it varies.}
  \label{tab:metrics}
  \scriptsize
  \setlength{\tabcolsep}{3pt}
  \resizebox{\columnwidth}{!}{%
  \begin{tabular}{@{}llrll@{}}
    \toprule
    \multicolumn{1}{>{\columncolor{bbllava}}l}{\textbf{Benchmark}} &
    \multicolumn{1}{>{\columncolor{bbllava}}l}{\textbf{Split}} &
    \multicolumn{1}{>{\columncolor{bbllava}}r}{\textbf{\#Q}} &
    \multicolumn{1}{>{\columncolor{bbllava}}l}{\textbf{\#Img}} &
    \multicolumn{1}{>{\columncolor{bbllava}}l}{\textbf{Metric}} \\
    \midrule

    \rowcolor{sectshade}
    \multicolumn{5}{l}{\textbf{\textit{Multi-image}}} \\
    Mantis-Eval \cite{jiang2024mantisinterleavedmultiimageinstruction}     & test     & 217     & 2--5 (2.49)  & Acc.\ (letter / short answer) \\
    NLVR2  \cite{suhr-etal-2019-corpus}       & test-P   & 6{,}967 & 2            & Acc.\ (letter) \\
    MuirBench \cite{wang2024muirbenchcomprehensivebenchmarkrobust}       & test     & 2{,}600 & 2--9 (4.33)  & Acc.\ (letter) \\
    BLINK  \cite{Fu_2024}          & val      & 1{,}901 & 1--4 (1.93)  & Acc.\ (letter) \\
    Q-Bench2   \cite{Zhang_2024}      & dev      & 1{,}000 & 2            & Acc.\ (letter) \\

    \midrule
    \rowcolor{sectshade}
    \multicolumn{5}{l}{\textbf{\textit{Single-image}}} \\
    TextVQA  \cite{Singh_2019_CVPR}        & val      & 5{,}000 & 1            & VQA accuracy \\
    ScienceQA-IMG \cite{NEURIPS2022_11332b6b}   & test     & 2{,}017 & 1            & Acc.\ (letter) \\
    MMStar  \cite{NEURIPS2024_2f8ee6a3}         & val      & 1{,}500 & 1            & Acc.\ (letter) \\
    POPE   \cite{li-etal-2023-evaluating}          & 3 splits & 9{,}000 & 1            & Acc., mean of splits \\

    \midrule
    \rowcolor{sectshade}
    \multicolumn{5}{l}{\textbf{\textit{Video}}} \\
    Video-MME \cite{fu2025video}       & test     & 2{,}700 & 32 frames    & Acc.\ (letter) \\
    EgoSchema    \cite{mangalam2023egoschema}    & labeled subset & 500 & 32 frames  & Acc.\ (letter) \\

    \midrule
    \rowcolor{sectshade}
    \multicolumn{5}{l}{\textbf{\textit{Embodied application}}} \\
    ERQA    \cite{geminiroboticsteam2025geminiroboticsbringingai}         & test     & 400     & 1--16 (1.57) & Acc.\ (letter) \\
    MV-RoboBench  \cite{feng2026seeingviewsbenchmarkingspatial}   & test     & 1{,}147 & 3--6 (3.67)  & Acc.\ (letter) \\
    OpenEQA   \cite{Majumdar_2024_CVPR}       & EM-EQA   & 1{,}636 & 16 frames    & LLM judge \\
    RoboSpatial  \cite{song2025robospatial}    & Home, yes/no & 228 & 1            & Acc., macro over types \\
    NuScenes-QA  \cite{qian2023nuscenes}    & mini val & 2{,}888 & 6 views      & Acc., macro over 5 types \\
    \bottomrule
  \end{tabular}%
  }
\end{table}

\subsection{Efficiency measurement and compute estimate}
\label{sec:appendix-flops}

\noindent\textbf{Measured cost} (Table~4).
Table~4 uses LLaVA-OneVision-7B at $B=\tfrac18$ and averages the five multi-image benchmarks.
Each method runs alone on a single NVIDIA H100, one prompt at a time. FLOPs are the
language-model prefill cost computed from the sequence length $s_\ell$ that each layer actually
processes, text included, as $\sum_\ell\big(c_{\mathrm{lin}}\,s_\ell+c_{\mathrm{attn}}\,s_\ell^{2}/2\big)$
with the constants defined below. Storage is the key-value cache size.
At $B=\tfrac18$, all methods realize similar budgets. AVTP, VisionZip, FastV, and
PyramidDrop are faster ($46.6$--$52.0$\,ms vs.\ $53.5$\,ms), reflecting the additional scoring and merging operations performed by
\method at $S(N)$ stages.

\begin{figure*}[!tp]
  \centering
  \includegraphics[width=\linewidth]{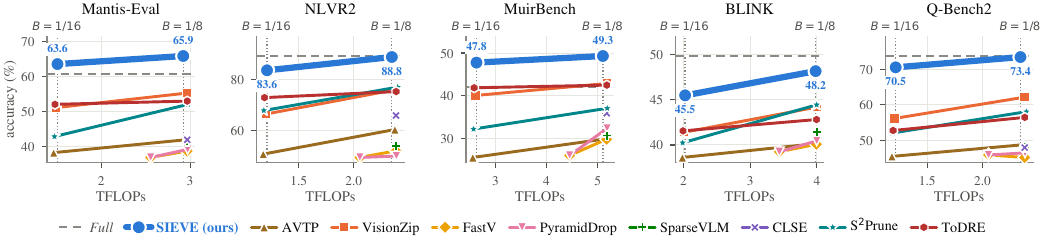}
  \caption{\textbf{Performance-efficiency trade-off curves} by benchmark on LLaVA-OneVision at $B=\tfrac1{16}$ and $\tfrac18$ (x-axis: estimated visual-token prefill TFLOPs). \method leads at both evaluated budgets on all benchmarks. FastV and PyramidDrop exceed the $\tfrac1{16}$ target even when retaining one token, so their left points lie right of the $B=\tfrac1{16}$ line. SparseVLM and CLSE are feasible only at $B=\tfrac18$.}
  \label{fig:flops-perds}
\end{figure*}

\noindent\textbf{Overshoot at $B=\tfrac1{16}$.}
FastV and PyramidDrop exceed the budget even when retaining only one token,
because the cost of their three full-width layers already exceeds the target. On NLVR2,
this prefix alone averages $3n/L=156.2$ visual tokens against a target of
$91.1$ (realized: $157.1$). Across the five benchmarks, their computation
exceeds that of \method by $46\%$ (\cref{fig:flops-perds}).

\noindent\textbf{Compute estimate} (Fig.~2).
Fig.~2 counts visual tokens only. We estimate visual-token prefill cost as
$L\big(c_{\mathrm{lin}}\,\bar n+c_{\mathrm{attn}}\,\bar n^{2}/2\big)$,
where $c_{\mathrm{lin}}$ is the per-token projection/MLP cost and
$c_{\mathrm{attn}}$ the per-pair attention cost of the 7B decoder. Text
tokens are excluded. The linear term uses the measured $\bar n$ (Eq.~(1))
and is exact. The quadratic term uses $\bar n^{2}$ in place of the layer
average of $n_\ell^{2}$, which slightly underestimates it for methods that
reduce tokens, and is under $4\%$ of the total at every evaluated budget.
\Cref{fig:flops-perds} gives per-benchmark results, including the
concurrent baseline AVTP~\cite{zhang2026multiimagevisualtokenpruning}.

\section{\texorpdfstring{Additional quantitative results}{Additional quantitative results}}
\label{sec:appendix-results}

Averaged over the four backbones, \method retains $99\%$ of the uncompressed
multi-image average at $B=\tfrac14$ ($67.00$ vs.\ $67.54$).

\subsection{Single-image results}
\label{sec:appendix-single}

\Cref{tab:single} reports the per-benchmark single-image results on all four backbones;
Table~3 reports their means. \method has the highest four-backbone average at both budgets, but not the highest average on every backbone: VisionZip is higher on Qwen2.5-VL at both budgets and on Qwen3-VL at $B=\tfrac18$, CLSE is higher on Qwen2.5-VL at $B=\tfrac18$, and VisionZip ties \method on LLaVA-OneVision at $B=\tfrac18$. AVTP and ToDRE are not evaluated on the single-image benchmarks.

\begin{table*}[!htbp]
  \centering
  \caption{\textbf{Performance comparison for single-image benchmarks} on LLaVA-OneVision-7B (\textbf{OV}), Qwen2.5-VL-7B (\textbf{Qwen2.5}), Qwen3-VL-8B (\textbf{Qwen3}) and InternVL3-8B (\textbf{IVL3}). SQA-I: ScienceQA-IMG. Per-benchmark scores and their average; the best average for each backbone and budget is \textbf{bold} and the second best \underline{underlined}. A dash marks a method that cannot meet the budget. $\ddagger$: exceeds the target budget even when retaining only one token after pruning.}
  \label{tab:single}
  \scriptsize
  \setlength{\tabcolsep}{1.6pt}
  \resizebox{\textwidth}{!}{%
  \begin{tabular}{@{}lcccccccccccccccc>{\columncolor{avgshade}}c>{\columncolor{avgshade}}c>{\columncolor{avgshade}}c>{\columncolor{avgshade}}c@{}}
    \toprule
    Method & \multicolumn{4}{c}{\textbf{TextVQA}} & \multicolumn{4}{c}{\textbf{SQA-I}} & \multicolumn{4}{c}{\textbf{MMStar}} & \multicolumn{4}{c}{\textbf{POPE}} & \multicolumn{4}{c}{\cellcolor{avgshade}\textbf{Avg}} \\
    \cmidrule(lr){2-5} \cmidrule(lr){6-9} \cmidrule(lr){10-13} \cmidrule(lr){14-17} \cmidrule(lr){18-21}
     & \cellcolor{bbllava}\textbf{OV} & \cellcolor{bbqwenv2}\textbf{Qwen2.5} & \cellcolor{bbqwenv3}\textbf{Qwen3} & \cellcolor{bbintern}\textbf{IVL3} & \cellcolor{bbllava}\textbf{OV} & \cellcolor{bbqwenv2}\textbf{Qwen2.5} & \cellcolor{bbqwenv3}\textbf{Qwen3} & \cellcolor{bbintern}\textbf{IVL3} & \cellcolor{bbllava}\textbf{OV} & \cellcolor{bbqwenv2}\textbf{Qwen2.5} & \cellcolor{bbqwenv3}\textbf{Qwen3} & \cellcolor{bbintern}\textbf{IVL3} & \cellcolor{bbllava}\textbf{OV} & \cellcolor{bbqwenv2}\textbf{Qwen2.5} & \cellcolor{bbqwenv3}\textbf{Qwen3} & \cellcolor{bbintern}\textbf{IVL3} & \cellcolor{bbllava}\textbf{OV} & \cellcolor{bbqwenv2}\textbf{Qwen2.5} & \cellcolor{bbqwenv3}\textbf{Qwen3} & \cellcolor{bbintern}\textbf{IVL3} \\
    \midrule
    \textit{Full} & \textit{76.21} & \textit{84.50} & \textit{83.53} & \textit{81.53} & \textit{95.93} & \textit{88.30} & \textit{94.55} & \textit{98.36} & \textit{61.33} & \textit{61.73} & \textit{60.13} & \textit{67.20} & \textit{89.22} & \textit{87.82} & \textit{88.82} & \textit{90.66} & \textit{80.67} & \textit{80.59} & \textit{81.76} & \textit{84.44} \\
    \midrule
    \rowcolor{sectshade} \multicolumn{21}{l}{\textbf{\textit{Keep 6.25\% of visual tokens}}} \\
    VisionZip~\cite{Yang_2025_CVPR} & 48.93 & 48.76 & 35.05 & 23.02 & 84.88 & 79.38 & 79.87 & 78.88 & 44.13 & 41.80 & 37.47 & 44.53 & 82.63 & 78.21 & 72.56 & 80.70 & \underline{65.14} & \textbf{62.04} & \underline{56.24} & 56.78 \\
    FastV~\cite{Chen_2024}$^{\ddagger}$ & 9.15 & 9.03 & 9.10 & 9.93 & 74.37 & 73.03 & 73.57 & 73.97 & 29.53 & 27.93 & 28.47 & 30.40 & 47.46 & 45.42 & 57.96 & 46.17 & 40.13 & 38.85 & 42.28 & 40.12 \\
    PyramidDrop~\cite{Xing_2025_CVPR}$^{\ddagger}$ & 9.15 & 9.03 & 9.10 & 9.93 & 74.37 & 73.03 & 73.57 & 73.97 & 29.53 & 27.93 & 28.47 & 30.40 & 47.46 & 45.42 & 57.96 & 46.17 & 40.13 & 38.85 & 42.28 & 40.12 \\
    CLSE~\cite{chen2026spectralevolutionguidedtokenpruning} & -- & -- & 9.66 & -- & -- & -- & 73.13 & -- & -- & -- & 29.07 & -- & -- & -- & 58.81 & -- & -- & -- & 42.67 & -- \\
    S$^2$Prune~\cite{jia2026s2prunespatiallystructuredvisual} & 35.14 & 42.02 & 15.98 & 35.76 & 81.26 & 75.58 & 77.79 & 80.81 & 39.53 & 34.96 & 33.93 & 41.13 & 70.58 & 67.08 & 73.48 & 79.03 & 56.63 & 54.91 & 50.30 & \underline{59.18} \\
    \cmidrule(lr){2-21}
    \rowcolor{ourshade} \textbf{\method} & 57.90 & 31.67 & 30.82 & 34.10 & 84.53 & 80.07 & 81.66 & 86.66 & 49.73 & 40.13 & 41.27 & 49.00 & 86.46 & 80.88 & 84.39 & 87.99 & \textbf{69.66} & \underline{58.19} & \textbf{59.53} & \textbf{64.44} \\
    \midrule
    \rowcolor{sectshade} \multicolumn{21}{l}{\textbf{\textit{Keep 12.5\% of visual tokens}}} \\
    VisionZip~\cite{Yang_2025_CVPR} & 61.98 & 62.28 & 58.92 & 42.84 & 88.60 & 83.44 & 84.18 & 86.71 & 50.33 & 46.33 & 42.67 & 49.60 & 86.96 & 82.64 & 82.51 & 86.40 & \textbf{71.97} & \textbf{68.67} & \textbf{67.07} & 66.39 \\
    FastV~\cite{Chen_2024} & 19.98 & 26.22 & 21.57 & 32.65 & 77.09 & 74.81 & 75.41 & 81.46 & 36.07 & 30.87 & 32.60 & 41.93 & 65.36 & 51.94 & 58.97 & 77.62 & 49.62 & 45.96 & 47.14 & 58.41 \\
    PyramidDrop~\cite{Xing_2025_CVPR} & 22.67 & 25.63 & 16.11 & 17.87 & 75.61 & 74.71 & 75.06 & 76.15 & 33.87 & 30.47 & 31.93 & 34.93 & 59.38 & 53.11 & 61.80 & 68.53 & 47.88 & 45.98 & 46.23 & 49.37 \\
    SparseVLM~\cite{pmlr-v267-zhang25s} & 23.63 & 42.87 & 41.27 & 43.13 & 78.98 & 78.78 & 79.77 & 86.17 & 39.47 & 35.27 & 38.00 & 47.80 & 69.19 & 69.14 & 76.30 & 84.77 & 52.82 & 56.52 & 58.84 & 65.47 \\
    CLSE~\cite{chen2026spectralevolutionguidedtokenpruning} & 34.94 & 67.77 & 37.66 & 53.25 & 81.51 & 80.66 & 77.34 & 92.27 & 44.53 & 44.20 & 38.53 & 54.27 & 78.38 & 75.60 & 75.06 & 87.14 & 59.84 & \underline{67.06} & 57.15 & \underline{71.73} \\
    S$^2$Prune~\cite{jia2026s2prunespatiallystructuredvisual} & 53.75 & 62.20 & 29.93 & 51.98 & 87.80 & 80.61 & 81.36 & 87.85 & 50.00 & 43.78 & 42.20 & 49.20 & 83.82 & 78.50 & 81.93 & 86.88 & \underline{68.84} & 66.27 & 58.86 & 68.98 \\
    \cmidrule(lr){2-21}
    \rowcolor{ourshade} \textbf{\method} & 62.95 & 49.03 & 46.88 & 61.15 & 86.71 & 81.85 & 83.64 & 93.21 & 51.20 & 45.60 & 46.27 & 57.60 & 87.02 & 84.34 & 86.01 & 90.00 & \textbf{71.97} & 65.21 & \underline{65.70} & \textbf{75.49} \\
    \bottomrule
  \end{tabular}%
  }
\end{table*}

\subsection{Prompts with many images}
\label{sec:appendix-n5}

On the 1{,}096 MuirBench prompts with $N\geq5$ (\cref{tab:n5}), averaged across the four backbones,
\method outperforms all baselines at every budget ($51.87$, $53.93$ and $54.47$ at $B=\tfrac1{16}$,
$\tfrac18$ and $\tfrac14$; the strongest baseline, ToDRE, reaches $50.05$, $52.42$ and $54.11$).
\method exceeds the uncompressed LLaVA-OneVision model at every evaluated budget.

\begin{table}[!htbp]
  \centering
  \caption{\textbf{Prompts with at least five images.} Accuracy on the $1096$ MuirBench prompts with $N \geq 5$, computed from the same predictions as Table~2. In each block, the best score in each column is \textbf{bold} and the second best \underline{underlined}. $\ddagger$: exceeds the target budget even when retaining only one token after pruning.}
  \label{tab:n5}
  \resizebox{\columnwidth}{!}{%
  \begin{tabular}{lcccc>{\columncolor{avgshade}}c}
    \toprule
    Method & \cellcolor{bbllava}\textbf{LLaVA-OneVision} & \cellcolor{bbqwenv2}\textbf{Qwen2.5-VL} & \cellcolor{bbqwenv3}\textbf{Qwen3-VL} & \cellcolor{bbintern}\textbf{InternVL3} & \textbf{Avg} \\
    \midrule
    \textit{Full} & \textit{47.90} & \textit{63.14} & \textit{52.65} & \textit{56.11} & \textit{54.95} \\
    \midrule
    \rowcolor{sectshade} \multicolumn{6}{l}{\textbf{\textit{Keep 6.25\% of visual tokens}}} \\
    VisionZip~\cite{Yang_2025_CVPR} & 40.88 & 48.27 & 50.18 & 41.88 & 45.30 \\
    FastV~\cite{Chen_2024}$^{\ddagger}$ & 26.00 & 29.47 & 46.44 & 18.98 & 30.22 \\
    PyramidDrop~\cite{Xing_2025_CVPR}$^{\ddagger}$ & 26.00 & 29.47 & 46.44 & 18.98 & 30.22 \\
    AVTP~\cite{zhang2026multiimagevisualtokenpruning} & 28.38 & 43.70 & 47.54 & 33.12 & 38.19 \\
    CLSE~\cite{chen2026spectralevolutionguidedtokenpruning} & --- & --- & 48.08 & --- & --- \\
    S$^2$Prune~\cite{jia2026s2prunespatiallystructuredvisual} & 37.23 & 48.72 & 47.26 & 40.15 & 43.34 \\
    ToDRE~\cite{li2026todre} & \underline{43.43} & \underline{52.28} & \textbf{52.28} & \textbf{52.19} & \underline{50.05} \\
    \cmidrule(lr){2-6}
    \rowcolor{ourshade} \textbf{\method} & \textbf{48.91} & \textbf{54.74} & \underline{51.92} & \underline{51.92} & \textbf{51.87} \\
    \midrule
    \rowcolor{sectshade} \multicolumn{6}{l}{\textbf{\textit{Keep 12.5\% of visual tokens}}} \\
    VisionZip~\cite{Yang_2025_CVPR} & \underline{45.44} & 55.66 & 49.09 & 53.38 & 50.89 \\
    FastV~\cite{Chen_2024} & 32.94 & 37.23 & 50.27 & 33.12 & 38.39 \\
    PyramidDrop~\cite{Xing_2025_CVPR} & 38.05 & 36.86 & \textbf{53.28} & 29.20 & 39.35 \\
    SparseVLM~\cite{pmlr-v267-zhang25s} & 34.31 & 41.06 & 49.73 & 35.58 & 40.17 \\
    AVTP~\cite{zhang2026multiimagevisualtokenpruning} & 32.76 & 48.81 & 49.36 & 43.16 & 43.52 \\
    CLSE~\cite{chen2026spectralevolutionguidedtokenpruning} & 37.14 & 49.82 & 48.54 & 40.51 & 44.00 \\
    S$^2$Prune~\cite{jia2026s2prunespatiallystructuredvisual} & 41.24 & 55.38 & 51.09 & 47.17 & 48.72 \\
    ToDRE~\cite{li2026todre} & 45.26 & \underline{56.48} & 51.55 & \textbf{56.39} & \underline{52.42} \\
    \cmidrule(lr){2-6}
    \rowcolor{ourshade} \textbf{\method} & \textbf{52.01} & \textbf{57.21} & \underline{51.92} & \underline{54.56} & \textbf{53.93} \\
    \midrule
    \rowcolor{sectshade} \multicolumn{6}{l}{\textbf{\textit{Keep 25\% of visual tokens}}} \\
    VisionZip~\cite{Yang_2025_CVPR} & 46.62 & 58.94 & 50.64 & 56.11 & 53.08 \\
    FastV~\cite{Chen_2024} & 37.96 & 51.46 & 50.64 & 50.46 & 47.63 \\
    PyramidDrop~\cite{Xing_2025_CVPR} & 40.33 & 51.82 & 51.82 & 48.72 & 48.17 \\
    SparseVLM~\cite{pmlr-v267-zhang25s} & 37.23 & 51.19 & 50.91 & 50.09 & 47.36 \\
    AVTP~\cite{zhang2026multiimagevisualtokenpruning} & 41.88 & 56.02 & \textbf{52.74} & 49.64 & 50.07 \\
    CLSE~\cite{chen2026spectralevolutionguidedtokenpruning} & 45.44 & 57.57 & 50.73 & 51.09 & 51.21 \\
    S$^2$Prune~\cite{jia2026s2prunespatiallystructuredvisual} & 46.72 & \textbf{60.58} & 51.82 & 52.19 & 52.83 \\
    ToDRE~\cite{li2026todre} & \underline{47.35} & 59.76 & \underline{52.10} & \textbf{57.21} & \underline{54.11} \\
    \cmidrule(lr){2-6}
    \rowcolor{ourshade} \textbf{\method} & \textbf{50.00} & \underline{60.22} & 51.00 & \underline{56.66} & \textbf{54.47} \\
    \bottomrule
\end{tabular}%
  }
\end{table}
\subsection{Video results by benchmark}
\label{sec:appendix-video}

\Cref{tab:video-full} separates the two benchmarks averaged in
Table~3, Video-MME (V) and EgoSchema (E). \method has the highest average at both budgets,
and its lead comes from EgoSchema, where it has the best score on Qwen2.5-VL and Qwen3-VL at both
budgets. On Video-MME, \method is best only on LLaVA-OneVision at $B=\tfrac1{16}$; ToDRE or VisionZip
is best in the other cells.

\begin{table}[!tp]
  \centering
  \caption{\textbf{Performance comparison for video benchmarks.} Each clip uses $32$ frames selected and encoded by the backbone's native video pipeline. \textbf{V}: Video-MME, all 2,700 questions; \textbf{E}: EgoSchema's 500-question labeled subset (0.2 points per question; the remaining 4,531 questions require leaderboard evaluation). \textbf{Avg} is the average of Table~3. The best score in each column is \textbf{bold} and the second best \underline{underlined}; a dash marks a method that cannot meet the budget. $\ddagger$: exceeds the target budget even when retaining only one token after pruning.}
  \label{tab:video-full}
  \scriptsize
  \setlength{\tabcolsep}{2.5pt}
  \resizebox{\columnwidth}{!}{%
  \begin{tabular}{@{}lcccccccc>{\columncolor{avgshade}}c@{}}
    \toprule
    & \multicolumn{2}{c}{\cellcolor{bbllava}\textbf{LLaVA-OneVision-7B}} & \multicolumn{2}{c}{\cellcolor{bbqwenv2}\textbf{Qwen2.5-VL-7B}} & \multicolumn{2}{c}{\cellcolor{bbqwenv3}\textbf{Qwen3-VL-8B}} & \multicolumn{2}{c}{\cellcolor{bbintern}\textbf{InternVL3-8B}} & \\
    \cmidrule(lr){2-3} \cmidrule(lr){4-5} \cmidrule(lr){6-7} \cmidrule(lr){8-9}
    Method & V & E & V & E & V & E & V & E & \textbf{Avg} \\
    \midrule
    \textit{Full} & \textit{58.44} & \textit{62.60} & \textit{62.19} & \textit{59.40} & \textit{64.26} & \textit{68.60} & \textit{65.56} & \textit{70.00} & \textit{63.88} \\
    \midrule
    \rowcolor{sectshade} \multicolumn{10}{l}{\textbf{\textit{Keep 6.25\% of visual tokens}}} \\
    VisionZip~\cite{Yang_2025_CVPR} & 49.78 & 55.00 & \textbf{56.52} & 51.20 & 55.56 & 59.40 & 55.85 & 54.60 & 54.74 \\
    FastV~\cite{Chen_2024}$^{\ddagger}$ & 39.07 & 36.20 & 40.33 & 30.40 & 41.74 & 37.60 & 41.52 & 36.00 & 37.86 \\
    PyramidDrop~\cite{Xing_2025_CVPR}$^{\ddagger}$ & 39.07 & 36.20 & 40.33 & 30.40 & 41.89 & 37.60 & 41.52 & 36.00 & 37.88 \\
    AVTP~\cite{zhang2026multiimagevisualtokenpruning} & 40.19 & 37.80 & 48.63 & 48.40 & 43.33 & 50.40 & 48.56 & 43.60 & 45.12 \\
    CLSE~\cite{chen2026spectralevolutionguidedtokenpruning} & -- & -- & -- & -- & 42.63 & 38.00 & -- & -- & -- \\
    S$^2$Prune~\cite{jia2026s2prunespatiallystructuredvisual} & 45.85 & 47.60 & 50.52 & 48.00 & 49.85 & 57.60 & 50.33 & 48.80 & 49.82 \\
    ToDRE~\cite{li2026todre} & \underline{51.22} & \textbf{63.60} & \underline{54.22} & \underline{61.40} & \textbf{56.44} & \underline{62.20} & \textbf{58.52} & \textbf{64.20} & \underline{58.97} \\
    \cmidrule(lr){2-10}
    \rowcolor{ourshade} \textbf{\method} & \textbf{54.11} & \underline{61.80} & 53.44 & \textbf{63.60} & \underline{55.81} & \textbf{68.60} & \underline{56.93} & \underline{63.80} & \textbf{59.76} \\
    \midrule
    \rowcolor{sectshade} \multicolumn{10}{l}{\textbf{\textit{Keep 12.5\% of visual tokens}}} \\
    VisionZip~\cite{Yang_2025_CVPR} & \textbf{55.15} & \underline{62.00} & \textbf{59.44} & 53.40 & \textbf{59.96} & \underline{65.60} & \textbf{60.00} & 62.80 & 59.79 \\
    FastV~\cite{Chen_2024} & 44.26 & 49.40 & 46.56 & 48.40 & 51.48 & 55.20 & 50.00 & 51.20 & 49.56 \\
    PyramidDrop~\cite{Xing_2025_CVPR} & 42.48 & 44.80 & 45.63 & 45.20 & 49.52 & 48.40 & 48.96 & 40.00 & 45.63 \\
    SparseVLM~\cite{pmlr-v267-zhang25s} & 44.26 & 53.40 & 51.19 & 59.20 & 56.22 & 64.80 & 53.56 & 59.80 & 55.31 \\
    AVTP~\cite{zhang2026multiimagevisualtokenpruning} & 43.74 & 46.80 & 51.37 & 57.00 & 49.56 & 58.60 & 51.30 & 53.80 & 51.52 \\
    CLSE~\cite{chen2026spectralevolutionguidedtokenpruning} & 47.89 & 58.80 & 51.78 & 61.00 & 53.15 & 62.00 & 55.59 & 64.40 & 56.83 \\
    S$^2$Prune~\cite{jia2026s2prunespatiallystructuredvisual} & 51.30 & 61.20 & 53.56 & 58.60 & 56.11 & 64.20 & 55.15 & 60.60 & 57.59 \\
    ToDRE~\cite{li2026todre} & 52.74 & \textbf{64.20} & \underline{56.00} & \underline{65.00} & \underline{58.26} & 64.00 & \underline{59.00} & \textbf{66.40} & \underline{60.70} \\
    \cmidrule(lr){2-10}
    \rowcolor{ourshade} \textbf{\method} & \underline{53.85} & \underline{62.00} & 53.89 & \textbf{67.40} & 57.96 & \textbf{71.00} & 57.93 & \underline{65.60} & \textbf{61.21} \\
    \bottomrule
  \end{tabular}%
  }
\end{table}
\subsection{Embodied application results by benchmark}
\label{sec:appendix-embodied}

\Cref{tab:robotics-full} separates the five benchmarks averaged in the embodied columns of
Table~3: ERQA (E), MV-RoboBench (M), OpenEQA (O), RoboSpatial (R; Home split) and
NuScenes-QA (N). \method has the highest average across backbones at both budgets.

\begin{table*}[t]
  \centering
  \caption{\textbf{Performance comparison for embodied application benchmarks.} The per-benchmark scores behind the embodied columns of Table~3. \textbf{E}: ERQA; \textbf{M}: MV-RoboBench; \textbf{O}: OpenEQA; \textbf{R}: RoboSpatial (Home split); \textbf{N}: NuScenes-QA. Avg is the unweighted mean of the five. Mean averages the four backbones' Avg. In each block, the best Avg for each backbone and the best Mean are \textbf{bold} and the second best \underline{underlined}; \textit{Full} is the model with no compression. A dash marks a method that cannot run at that budget. $^{\ddagger}$: exceeds the target budget even when retaining only one token after pruning.}
  \label{tab:robotics-full}
  \scriptsize
  \setlength{\tabcolsep}{1.0pt}
  \resizebox{\textwidth}{!}{%
  \begin{tabular}{@{}l*{4}{ccccc>{\columncolor{avgshade}}c}>{\columncolor{avgshade}}c@{}}
    \toprule
    & \multicolumn{6}{c}{\cellcolor{bbllava}\textbf{LLaVA-OneVision-7B}} & \multicolumn{6}{c}{\cellcolor{bbqwenv2}\textbf{Qwen2.5-VL-7B}} & \multicolumn{6}{c}{\cellcolor{bbqwenv3}\textbf{Qwen3-VL-8B}} & \multicolumn{6}{c}{\cellcolor{bbintern}\textbf{InternVL3-8B}} & \\
    \cmidrule(lr){2-7} \cmidrule(lr){8-13} \cmidrule(lr){14-19} \cmidrule(lr){20-25}
    Method & E & M & O & R & N & \textbf{Avg} & E & M & O & R & N & \textbf{Avg} & E & M & O & R & N & \textbf{Avg} & E & M & O & R & N & \textbf{Avg} & \textbf{Mean} \\
    \midrule
    \textit{Full} & \textit{37.75} & \textit{20.31} & \textit{51.47} & \textit{58.63} & \textit{32.59} & \textit{40.15} & \textit{40.25} & \textit{24.15} & \textit{46.01} & \textit{55.90} & \textit{29.21} & \textit{39.10} & \textit{43.00} & \textit{25.72} & \textit{54.00} & \textit{73.09} & \textit{32.52} & \textit{45.67} & \textit{38.50} & \textit{22.41} & \textit{52.69} & \textit{62.89} & \textit{34.55} & \textit{42.21} & \textit{41.78} \\
    \midrule
    \rowcolor{sectshade} \multicolumn{26}{l}{\textbf{\textit{Keep 6.25\% of visual tokens}}} \\
    VisionZip~\cite{Yang_2025_CVPR} & 33.75 & 19.79 & 42.22 & 58.37 & 32.44 & 37.31 & 36.50 & 21.36 & 43.57 & 46.95 & 28.58 & \underline{35.39} & 32.25 & 20.84 & 43.78 & 67.35 & 27.12 & 38.27 & 33.75 & 18.92 & 40.74 & 56.35 & 30.40 & 36.03 & 36.75 \\
    FastV~\cite{Chen_2024}$^{\ddagger}$ & 33.25 & 17.61 & 34.69 & 53.50 & 23.22 & 32.45 & 33.25 & 19.18 & 33.21 & 30.52 & 26.82 & 28.60 & 30.50 & 19.44 & 35.48 & 72.80 & 27.56 & 37.16 & 31.75 & 19.70 & 33.80 & 63.95 & 29.61 & 35.76 & 33.49 \\
    PyramidDrop~\cite{Xing_2025_CVPR}$^{\ddagger}$ & 33.25 & 17.61 & 34.69 & 53.50 & 23.26 & 32.46 & 33.25 & 19.18 & 33.21 & 30.52 & 27.08 & 28.65 & 30.50 & 19.44 & 35.48 & 72.80 & 27.46 & 37.14 & 31.75 & 19.70 & 33.73 & 63.95 & 29.61 & 35.75 & 33.50 \\
    AVTP~\cite{zhang2026multiimagevisualtokenpruning} & 31.25 & 17.79 & 33.62 & 55.11 & 21.59 & 31.87 & 31.00 & 19.35 & 37.39 & 33.58 & 27.20 & 29.70 & 29.50 & 20.66 & 35.61 & 73.07 & 28.47 & 37.46 & 31.50 & 19.09 & 35.31 & 53.65 & 28.79 & 33.67 & 33.18 \\
    CLSE~\cite{chen2026spectralevolutionguidedtokenpruning} & -- & -- & -- & -- & -- & -- & -- & -- & -- & -- & -- & -- & 31.75 & 20.40 & 34.57 & 67.38 & 27.10 & 36.24 & -- & -- & -- & -- & -- & -- & -- \\
    S$^2$Prune~\cite{jia2026s2prunespatiallystructuredvisual} & 32.75 & 18.48 & 40.66 & 53.82 & 25.28 & 34.20 & 34.42 & 20.58 & 42.13 & 40.43 & 27.62 & 33.04 & 35.00 & 20.66 & 44.24 & 69.19 & 31.31 & 40.08 & 31.75 & 19.35 & 40.19 & 51.11 & 32.35 & 34.95 & 35.57 \\
    ToDRE~\cite{li2026todre} & 34.50 & 18.66 & 45.29 & 62.44 & 30.94 & \underline{38.37} & 34.50 & 20.92 & 44.16 & 46.67 & 29.27 & 35.10 & 31.50 & 21.19 & 46.65 & 72.10 & 30.87 & \underline{40.46} & 35.50 & 21.62 & 46.85 & 63.04 & 34.40 & \textbf{40.28} & \underline{38.55} \\
    \cmidrule(lr){2-26}
    \rowcolor{ourshade} \textbf{\method} & 32.50 & 17.35 & 48.30 & 62.44 & 31.53 & \textbf{38.42} & 32.50 & 21.36 & 44.10 & 54.47 & 28.10 & \textbf{36.11} & 36.00 & 22.23 & 45.55 & 76.36 & 30.21 & \textbf{42.07} & 35.00 & 18.66 & 46.61 & 62.83 & 33.82 & \underline{39.38} & \textbf{39.00} \\
    
    \midrule
    \rowcolor{sectshade} \multicolumn{26}{l}{\textbf{\textit{Keep 12.5\% of visual tokens}}} \\
    VisionZip~\cite{Yang_2025_CVPR} & 32.75 & 19.53 & 47.22 & 65.10 & 33.35 & \underline{39.59} & 37.50 & 22.14 & 45.23 & 50.39 & 28.94 & \underline{36.84} & 37.00 & 23.54 & 48.26 & 68.21 & 29.46 & 41.29 & 33.25 & 20.05 & 45.48 & 64.54 & 32.27 & 39.12 & 39.21 \\
    FastV~\cite{Chen_2024} & 30.00 & 18.22 & 35.16 & 55.40 & 26.40 & 33.04 & 33.50 & 21.10 & 36.12 & 35.55 & 26.81 & 30.62 & 32.50 & 21.01 & 43.22 & 67.29 & 26.53 & 38.11 & 31.00 & 21.27 & 39.12 & 58.94 & 30.34 & 36.13 & 34.48 \\
    PyramidDrop~\cite{Xing_2025_CVPR} & 32.00 & 18.05 & 34.83 & 53.16 & 25.05 & 32.62 & 32.25 & 20.14 & 37.53 & 35.76 & 28.54 & 30.84 & 33.50 & 19.09 & 38.46 & 69.69 & 26.41 & 37.43 & 31.75 & 19.35 & 37.59 & 58.47 & 30.71 & 35.57 & 34.11 \\
    SparseVLM~\cite{pmlr-v267-zhang25s} & 31.25 & 17.35 & 35.38 & 53.64 & 26.45 & 32.81 & 32.50 & 19.27 & 38.94 & 39.15 & 26.39 & 31.25 & 33.25 & 21.62 & 47.07 & 71.21 & 28.32 & 40.29 & 39.25 & 19.62 & 42.16 & 61.30 & 31.86 & 38.84 & 35.80 \\
    AVTP~\cite{zhang2026multiimagevisualtokenpruning} & 32.25 & 17.79 & 37.09 & 49.95 & 26.13 & 32.64 & 32.25 & 20.14 & 38.81 & 41.25 & 29.13 & 32.32 & 32.50 & 20.66 & 36.51 & 68.72 & 29.05 & 37.49 & 30.25 & 18.74 & 38.88 & 51.39 & 32.20 & 34.29 & 34.19 \\
    CLSE~\cite{chen2026spectralevolutionguidedtokenpruning} & 32.25 & 20.05 & 41.79 & 52.20 & 28.20 & 34.90 & 33.00 & 21.62 & 39.82 & 45.18 & 27.46 & 33.42 & 34.75 & 22.14 & 45.72 & 72.85 & 28.85 & 40.86 & 36.25 & 21.27 & 45.71 & 60.33 & 34.64 & 39.64 & 37.21 \\
    S$^2$Prune~\cite{jia2026s2prunespatiallystructuredvisual} & 35.00 & 18.74 & 46.65 & 57.82 & 31.30 & 37.90 & 34.25 & 22.49 & 44.91 & 50.27 & 28.76 & 36.14 & 35.75 & 21.62 & 48.79 & 71.07 & 31.98 & 41.84 & 34.75 & 19.88 & 47.10 & 60.20 & 32.95 & 38.98 & 38.71 \\
    ToDRE~\cite{li2026todre} & 34.25 & 18.57 & 46.39 & 61.08 & 31.81 & 38.42 & 34.25 & 22.14 & 44.89 & 50.46 & 29.24 & 36.20 & 33.00 & 20.92 & 49.13 & 75.20 & 31.16 & \underline{41.88} & 36.50 & 22.49 & 49.79 & 64.39 & 34.31 & \textbf{41.50} & \underline{39.50} \\
    \cmidrule(lr){2-26}
    \rowcolor{ourshade} \textbf{\method} & 36.00 & 18.48 & 50.86 & 63.67 & 32.06 & \textbf{40.21} & 35.00 & 21.53 & 44.94 & 58.48 & 29.45 & \textbf{37.88} & 36.00 & 21.71 & 49.07 & 74.80 & 31.32 & \textbf{42.58} & 35.25 & 19.88 & 48.32 & 63.51 & 35.53 & \underline{40.50} & \textbf{40.29} \\

    \bottomrule
  \end{tabular}%
  }
\end{table*}
\subsection{Per-backbone ablations}
\label{sec:appendix-abl-onevision}
\label{sec:appendix-abl-backbones}

\Cref{tab:ablation-onevision-full,tab:ablation-qwen,tab:ablation-qwen3,tab:ablation-internvl}
break down the Table~6 ablations by backbone. Removing pooling causes the largest accuracy
drop on each backbone. Adding FastV or PyramidDrop after pooling improves on pooling alone
on Qwen3-VL and InternVL3 and reduces accuracy on LLaVA-OneVision and Qwen2.5-VL.

\begin{table}[!htbp]
  \centering
  \caption{\textbf{Ablations on LLaVA-OneVision-7B} at $B=\tfrac1{16}$. Each row changes one component of \method, with the same variants and grouping as Table~6. $\boldsymbol{\Delta}$ is the change from \method.}
  \label{tab:ablation-onevision-full}
  \scriptsize
  \setlength{\tabcolsep}{2.1pt}
  \renewcommand{\arraystretch}{0.92}
  \resizebox{\columnwidth}{!}{%
  \begin{tabular}{@{}lccccc>{\columncolor{avgshade}}rr@{}}
    \toprule
    & \cellcolor{bbllava}\textbf{Mantis} & \cellcolor{bbllava}\textbf{NLVR2} & \cellcolor{bbllava}\textbf{Muir} & \cellcolor{bbllava}\textbf{BLINK} & \cellcolor{bbllava}\textbf{Q-Bench2} & \textbf{Avg} & $\boldsymbol{\Delta}$ \\
    \midrule
    \rowcolor{ourshade} \textbf{\method} & \textbf{63.59} & \textbf{83.61} & \textbf{47.77} & \textbf{45.50} & \textbf{70.50} & \textbf{62.19} & \\
    \midrule
    \rowcolor{sectshade} \multicolumn{8}{@{}l}{\textbf{RQ1: Cross-image coverage and prompt-guided selection}} \\
    global top-$k$ (no per-image quota) & 59.91 & 82.17 & 47.96 & 40.93 & 69.60 & 60.11 & \dbad{$-2.08$} \\
    random selection scores & 60.37 & 83.67 & 49.27 & 40.93 & 67.40 & 60.33 & \dbad{$-1.86$} \\
    \midrule
    \rowcolor{sectshade} \multicolumn{8}{@{}l}{\textbf{RQ2: Token aggregation and image-identity preservation}} \\
    drop unselected tokens & 60.83 & 83.28 & 48.31 & 42.08 & 69.20 & 60.74 & \dbad{$-1.45$} \\
    allow cross-image merge targets & 63.13 & 83.62 & 48.08 & 40.98 & 69.70 & 61.10 & \dbad{$-1.09$} \\
    spatially nearest same-image target & 63.13 & 82.68 & 48.08 & 41.08 & 68.70 & 60.73 & \dbad{$-1.46$} \\
    \midrule
    \rowcolor{sectshade} \multicolumn{8}{@{}l}{\textbf{RQ3: Initial pooling and progressive reduction}} \\
    no pooling (selection at layer 0) & 45.62 & 62.01 & 33.27 & 41.35 & 50.30 & 46.51 & \dbad{$-15.68$} \\
    pooling only & 61.29 & 80.87 & 47.12 & 40.45 & 62.70 & 58.49 & \dbad{$-3.70$} \\
    pooling + FastV & 57.60 & 78.21 & 42.62 & 43.92 & 68.10 & 58.09 & \dbad{$-4.10$} \\
    pooling + PyramidDrop & 58.06 & 76.65 & 41.81 & 44.40 & 71.20 & 58.42 & \dbad{$-3.77$} \\
    \midrule
    \rowcolor{sectshade} \multicolumn{8}{@{}l}{\textbf{RQ4: Depth-wise placement of reduction stages}} \\
    first stage at layer 0 & 62.21 & 82.52 & 48.62 & 42.66 & 68.60 & 60.92 & \dbad{$-1.27$} \\
    first stage at layer 8 & 58.06 & 77.16 & 45.19 & 40.87 & 70.80 & 58.42 & \dbad{$-3.77$} \\
    later stages evenly spaced & 60.37 & 84.58 & 45.96 & 41.71 & 70.00 & 60.52 & \dbad{$-1.67$} \\
    later stages in the first half & 58.53 & 80.02 & 40.38 & 40.45 & 70.60 & 58.00 & \dbad{$-4.19$} \\
    \bottomrule
  \end{tabular}%
  }
\end{table}
\begin{table}[!htbp]
  \centering
  \caption{\textbf{Ablations on Qwen2.5-VL-7B} at $B=\tfrac1{16}$. Each row changes one component of \method, with the same variants and grouping as Table~6. $\boldsymbol{\Delta}$ is the change from \method.}
  \label{tab:ablation-qwen}
  \scriptsize
  \setlength{\tabcolsep}{2.1pt}
  \renewcommand{\arraystretch}{0.92}
  \resizebox{\columnwidth}{!}{%
  \begin{tabular}{@{}lccccc>{\columncolor{avgshade}}rr@{}}
    \toprule
    & \cellcolor{bbqwenv2}\textbf{Mantis} & \cellcolor{bbqwenv2}\textbf{NLVR2} & \cellcolor{bbqwenv2}\textbf{Muir} & \cellcolor{bbqwenv2}\textbf{BLINK} & \cellcolor{bbqwenv2}\textbf{Q-Bench2} & \textbf{Avg} & $\boldsymbol{\Delta}$ \\
    \midrule
    \rowcolor{ourshade} \textbf{\method} & \textbf{71.43} & \textbf{78.05} & \textbf{50.96} & \textbf{49.50} & \textbf{68.40} & \textbf{63.67} & \\
    \midrule
    \rowcolor{sectshade} \multicolumn{8}{@{}l}{\textbf{RQ1: Cross-image coverage and prompt-guided selection}} \\
    global top-$k$ (no per-image quota) & 67.28 & 77.84 & 45.65 & 48.92 & 65.60 & 61.06 & \dbad{$-2.61$} \\
    random selection scores & 67.74 & 77.39 & 45.15 & 47.45 & 65.80 & 60.71 & \dbad{$-2.96$} \\
    \midrule
    \rowcolor{sectshade} \multicolumn{8}{@{}l}{\textbf{RQ2: Token aggregation and image-identity preservation}} \\
    drop unselected tokens & 70.97 & 77.57 & 45.50 & 48.55 & 66.20 & 61.76 & \dbad{$-1.91$} \\
    allow cross-image merge targets & 69.59 & 77.88 & 45.88 & 49.40 & 65.30 & 61.61 & \dbad{$-2.06$} \\
    spatially nearest same-image target & 69.59 & 77.72 & 45.38 & 48.76 & 64.30 & 61.15 & \dbad{$-2.52$} \\
    \midrule
    \rowcolor{sectshade} \multicolumn{8}{@{}l}{\textbf{RQ3: Initial pooling and progressive reduction}} \\
    no pooling (selection at layer 0) & 60.37 & 66.28 & 42.00 & 46.24 & 65.80 & 56.14 & \dbad{$-7.53$} \\
    pooling only & 68.20 & 76.42 & 42.81 & 47.13 & 64.30 & 59.77 & \dbad{$-3.90$} \\
    pooling + FastV & 65.44 & 73.27 & 43.23 & 48.19 & 64.70 & 58.97 & \dbad{$-4.70$} \\
    pooling + PyramidDrop & 65.44 & 71.81 & 42.31 & 47.97 & 66.30 & 58.77 & \dbad{$-4.90$} \\
    \midrule
    \rowcolor{sectshade} \multicolumn{8}{@{}l}{\textbf{RQ4: Depth-wise placement of reduction stages}} \\
    first stage at layer 0 & 71.43 & 78.94 & 46.08 & 48.50 & 65.90 & 62.17 & \dbad{$-1.50$} \\
    first stage at layer 8 & 64.06 & 71.90 & 42.96 & 46.82 & 65.80 & 58.31 & \dbad{$-5.36$} \\
    later stages evenly spaced & 71.43 & 77.87 & 44.35 & 49.03 & 66.30 & 61.80 & \dbad{$-1.87$} \\
    later stages in the first half & 66.36 & 74.22 & 40.92 & 46.98 & 65.60 & 58.82 & \dbad{$-4.85$} \\
    \bottomrule
  \end{tabular}%
  }
\end{table}
\begin{table}[!htbp]
  \centering
  \caption{\textbf{Ablations on Qwen3-VL-8B} at $B=\tfrac1{16}$. Each row changes one component of \method, with the same variants and grouping as Table~6. $\boldsymbol{\Delta}$ is the change from \method.}
  \label{tab:ablation-qwen3}
  \scriptsize
  \setlength{\tabcolsep}{2.1pt}
  \renewcommand{\arraystretch}{0.92}
  \resizebox{\columnwidth}{!}{%
  \begin{tabular}{@{}lccccc>{\columncolor{avgshade}}rr@{}}
    \toprule
    & \cellcolor{bbqwenv3}\textbf{Mantis} & \cellcolor{bbqwenv3}\textbf{NLVR2} & \cellcolor{bbqwenv3}\textbf{Muir} & \cellcolor{bbqwenv3}\textbf{BLINK} & \cellcolor{bbqwenv3}\textbf{Q-Bench2} & \textbf{Avg} & $\boldsymbol{\Delta}$ \\
    \midrule
    \rowcolor{ourshade} \textbf{\method} & \textbf{60.83} & \textbf{73.20} & \textbf{47.27} & \textbf{44.29} & \textbf{68.50} & \textbf{58.82} & \\
    \midrule
    \rowcolor{sectshade} \multicolumn{8}{@{}l}{\textbf{RQ1: Cross-image coverage and prompt-guided selection}} \\
    global top-$k$ (no per-image quota) & 59.45 & 72.05 & 43.81 & 44.29 & 67.30 & 57.38 & \dbad{$-1.44$} \\
    random selection scores & 58.99 & 72.48 & 43.96 & 41.45 & 64.40 & 56.26 & \dbad{$-2.56$} \\
    \midrule
    \rowcolor{sectshade} \multicolumn{8}{@{}l}{\textbf{RQ2: Token aggregation and image-identity preservation}} \\
    drop unselected tokens & 60.37 & 73.10 & 44.12 & 44.71 & 67.80 & 58.02 & \dbad{$-0.80$} \\
    allow cross-image merge targets & 59.45 & 73.42 & 43.69 & 44.45 & 68.30 & 57.86 & \dbad{$-0.96$} \\
    spatially nearest same-image target & 58.53 & 72.97 & 43.46 & 44.29 & 67.60 & 57.37 & \dbad{$-1.45$} \\
    \midrule
    \rowcolor{sectshade} \multicolumn{8}{@{}l}{\textbf{RQ3: Initial pooling and progressive reduction}} \\
    no pooling (selection at layer 0) & 53.46 & 60.04 & 42.54 & 43.08 & 69.90 & 53.80 & \dbad{$-5.02$} \\
    pooling only & 58.99 & 70.81 & 42.12 & 42.82 & 65.00 & 55.95 & \dbad{$-2.87$} \\
    pooling + FastV & 57.60 & 67.17 & 45.15 & 44.19 & 67.90 & 56.40 & \dbad{$-2.42$} \\
    pooling + PyramidDrop & 62.21 & 67.49 & 44.96 & 45.82 & 67.60 & 57.62 & \dbad{$-1.20$} \\
    \midrule
    \rowcolor{sectshade} \multicolumn{8}{@{}l}{\textbf{RQ4: Depth-wise placement of reduction stages}} \\
    first stage at layer 0 & 59.45 & 72.70 & 44.69 & 43.71 & 67.40 & 57.59 & \dbad{$-1.23$} \\
    first stage at layer 8 & 58.53 & 69.18 & 44.81 & 43.61 & 66.90 & 56.61 & \dbad{$-2.21$} \\
    later stages evenly spaced & 59.45 & 74.32 & 45.04 & 45.13 & 68.00 & 58.39 & \dbad{$-0.43$} \\
    later stages in the first half & 57.60 & 70.39 & 45.50 & 43.77 & 67.70 & 56.99 & \dbad{$-1.83$} \\
    \bottomrule
  \end{tabular}%
  }
\end{table}
\begin{table}[!htbp]
  \centering
  \caption{\textbf{Ablations on InternVL3-8B} at $B=\tfrac1{16}$. Each row changes one component of \method, with the same variants and grouping as Table~6. $\boldsymbol{\Delta}$ is the change from \method.}
  \label{tab:ablation-internvl}
  \scriptsize
  \setlength{\tabcolsep}{2.1pt}
  \renewcommand{\arraystretch}{0.92}
  \resizebox{\columnwidth}{!}{%
  \begin{tabular}{@{}lccccc>{\columncolor{avgshade}}rr@{}}
    \toprule
    & \cellcolor{bbintern}\textbf{Mantis} & \cellcolor{bbintern}\textbf{NLVR2} & \cellcolor{bbintern}\textbf{Muir} & \cellcolor{bbintern}\textbf{BLINK} & \cellcolor{bbintern}\textbf{Q-Bench2} & \textbf{Avg} & $\boldsymbol{\Delta}$ \\
    \midrule
    \rowcolor{ourshade} \textbf{\method} & \textbf{67.74} & \textbf{78.30} & \textbf{50.92} & \textbf{49.50} & \textbf{71.40} & \textbf{63.57} & \\
    \midrule
    \rowcolor{sectshade} \multicolumn{8}{@{}l}{\textbf{RQ1: Cross-image coverage and prompt-guided selection}} \\
    global top-$k$ (no per-image quota) & 64.52 & 75.69 & 47.19 & 47.34 & 65.40 & 60.03 & \dbad{$-3.54$} \\
    random selection scores & 59.91 & 73.26 & 45.88 & 47.24 & 63.00 & 57.86 & \dbad{$-5.71$} \\
    \midrule
    \rowcolor{sectshade} \multicolumn{8}{@{}l}{\textbf{RQ2: Token aggregation and image-identity preservation}} \\
    drop unselected tokens & 65.90 & 76.03 & 47.08 & 47.34 & 65.30 & 60.33 & \dbad{$-3.24$} \\
    allow cross-image merge targets & 65.90 & 75.90 & 46.73 & 47.24 & 64.70 & 60.09 & \dbad{$-3.48$} \\
    spatially nearest same-image target & 66.36 & 74.78 & 46.42 & 48.66 & 63.90 & 60.02 & \dbad{$-3.55$} \\
    \midrule
    \rowcolor{sectshade} \multicolumn{8}{@{}l}{\textbf{RQ3: Initial pooling and progressive reduction}} \\
    no pooling (selection at layer 0) & 60.37 & 69.79 & 42.27 & 45.82 & 69.60 & 57.57 & \dbad{$-6.00$} \\
    pooling only & 61.75 & 72.63 & 44.38 & 47.92 & 64.10 & 58.16 & \dbad{$-5.41$} \\
    pooling + FastV & 64.98 & 76.19 & 45.46 & 48.76 & 64.70 & 60.02 & \dbad{$-3.55$} \\
    pooling + PyramidDrop & 67.28 & 74.49 & 45.19 & 50.08 & 67.00 & 60.81 & \dbad{$-2.76$} \\
    \midrule
    \rowcolor{sectshade} \multicolumn{8}{@{}l}{\textbf{RQ4: Depth-wise placement of reduction stages}} \\
    first stage at layer 0 & 64.06 & 76.04 & 47.00 & 47.92 & 64.00 & 59.80 & \dbad{$-3.77$} \\
    first stage at layer 8 & 65.90 & 72.79 & 44.12 & 46.92 & 63.20 & 58.59 & \dbad{$-4.98$} \\
    later stages evenly spaced & 66.36 & 76.10 & 44.27 & 47.08 & 65.50 & 59.86 & \dbad{$-3.71$} \\
    later stages in the first half & 66.82 & 74.59 & 41.23 & 46.98 & 64.10 & 58.74 & \dbad{$-4.83$} \\
    \bottomrule
  \end{tabular}%
  }
\end{table}

\subsection{Additional ablations}
\label{sec:appendix-abl-extra}

\Cref{tab:ablation-extra} reports additional variants at $B=\tfrac1{16}$.
LLaVA-OneVision encodes each image into $729$ tokens in our multi-image setup,
so allocating slots by token count is equivalent to equal allocation. The
\emph{floor + attention} variant in Table~5 gives each image $q_{\min}$ slots
and assigns the remainder in proportion to prompt-attention mass.

Randomizing scores only at the first stage causes nearly the same loss as randomizing
every stage on Qwen2.5-VL, Qwen3-VL, and InternVL3
(\cref{tab:ablation-qwen,tab:ablation-qwen3,tab:ablation-internvl}). This result
complements Fig.~1(a): early prompt attention helps rank tokens within each image
even when global attention rankings give the queried image only its uniform share.
Strided subsampling costs $7.59$ points on LLaVA-OneVision and at most $0.97$ points
on backbones whose encoders already combine neighboring patches.

\begin{table}[!htbp]
  \centering
  \caption{\textbf{Additional ablations} at $B=\tfrac1{16}$. Each entry is the change in the
  five-benchmark multi-image average from \method on that backbone (\method: $62.19$, $63.67$,
  $58.82$, $63.57$). A dash marks a variant not evaluated on that backbone; $=$ marks a variant that
  is identical to \method on that backbone.}
  \label{tab:ablation-extra}
  \scriptsize
  \setlength{\tabcolsep}{2.4pt}
  \renewcommand{\arraystretch}{0.92}
  \resizebox{\columnwidth}{!}{%
  \begin{tabular}{@{}lcccc@{}}
    \toprule
    Variant & \cellcolor{bbllava}\textbf{OneVision} & \cellcolor{bbqwenv2}\textbf{Qwen2.5} & \cellcolor{bbqwenv3}\textbf{Qwen3} & \cellcolor{bbintern}\textbf{InternVL3} \\
    \midrule
    \rowcolor{sectshade} \multicolumn{5}{@{}l}{\textbf{Allocation} \textcolor{black!60}{(\method: equal quota per image)}} \\
    floor + attention & \dbad{$-1.71$} & -- & -- & -- \\
    size-proportional share & $=$ & \dbad{$-2.10$} & \dbad{$-1.19$} & \dbad{$-3.34$} \\
    \midrule
    \rowcolor{sectshade} \multicolumn{5}{@{}l}{\textbf{Selection score} \textcolor{black!60}{(\method: prompt attention)}} \\
    random score, first stage only & \dbad{$-1.64$} & \dbad{$-2.99$} & \dbad{$-2.56$} & \dbad{$-5.70$} \\
    random score, later stages only & \dbad{$-0.66$} & \dbad{$-2.03$} & \dbad{$-0.77$} & \dbad{$-3.31$} \\
    \midrule
    \rowcolor{sectshade} \multicolumn{5}{@{}l}{\textbf{Merge target} \textcolor{black!60}{(\method: most similar same-image kept token)}} \\
    assign by attention weight & \dbad{$-1.64$} & \dbad{$-2.01$} & \dbad{$-1.59$} & \dbad{$-3.50$} \\
    random kept token & \dbad{$-2.37$} & \dbad{$-3.68$} & \dbad{$-1.03$} & \dbad{$-4.19$} \\
    \midrule
    \rowcolor{sectshade} \multicolumn{5}{@{}l}{\textbf{Pooling} \textcolor{black!60}{(\method: area-mean, $\alpha=3$)}} \\
    strided subsampling & \dbad{$-7.59$} & \dbad{$-0.97$} & \dbad{$-0.48$} & \dbad{$-0.72$} \\
    bilinear resampling & \dbad{$-1.21$} & -- & -- & -- \\
    less headroom, $\alpha=1.5$ & \dbad{$-3.15$} & -- & -- & -- \\
    \midrule
    \rowcolor{sectshade} \multicolumn{5}{@{}l}{\textbf{Position indices} \textcolor{black!60}{(\method: renumbered; a kept token keeps its index)}} \\
    average the positions of merged tokens & \dbad{$-1.36$} & -- & -- & -- \\
    keep pre-pooling indices & \dbad{$-2.95$} & -- & -- & \dbad{$-4.69$} \\
    \midrule
    \rowcolor{sectshade} \multicolumn{5}{@{}l}{\textbf{Schedule} \textcolor{black!60}{(\method: $\ell_1=2$, $S(N)$, geometric $\gamma=\tfrac12$)}} \\
    first stage at layer $4$ & \dbad{$-2.54$} & \dbad{$-2.48$} & \dbad{$-0.85$} & \dbad{$-3.86$} \\
    fixed $S=2$ & \dbad{$-1.18$} & \dbad{$-2.14$} & \dbad{$-0.81$} & \dbad{$-3.79$} \\
    fixed $S=3$ & \dbad{$-0.47$} & \dbad{$-1.62$} & \dbad{$-0.16$} & \dbad{$-3.10$} \\
    faster decay, $\gamma=\tfrac14$ & \dbad{$-0.53$} & -- & -- & -- \\
    slower decay, $\gamma=\tfrac34$ & \dbad{$-1.43$} & -- & -- & -- \\
    constant retention ratio & \dbad{$-1.03$} & \dbad{$-1.86$} & \dbad{$-0.80$} & \dbad{$-3.70$} \\
    \bottomrule
  \end{tabular}%
  }
\end{table}

\subsection{Progressive versus recursive reduction}
\label{sec:appendix-recursive}

\Cref{tab:recursive-single} complements the multi-image results in Table~7 with
single-image averages. The recursive variant selects and merges at a fixed layer $\ell$,
then restarts from the first layer. The budget includes the repeated scoring passes.
Infeasible Qwen2.5-VL recursive configurations exclude up to 70 of 1,500 MMStar,
15 of 2,017 ScienceQA-IMG, and 1 of 1,901 BLINK prompts per cell.

\begin{table}[!htbp]
  \centering
  \caption{\textbf{Progressive versus recursive reduction.} Mean accuracy over the
  four single-image benchmarks. Bold marks the highest average in each column.}
  \label{tab:recursive-single}
  \scriptsize
  \resizebox{\columnwidth}{!}{%
  \begin{tabular}{@{}lcccc@{}}
    \toprule
    & \multicolumn{2}{c}{\cellcolor{bbllava}\textbf{LLaVA-OneVision-7B}}
    & \multicolumn{2}{c}{\cellcolor{bbqwenv2}\textbf{Qwen2.5-VL-7B}} \\
    \cmidrule(lr){2-3}\cmidrule(lr){4-5}
    Method & 6.25\% & 12.5\% & 6.25\% & 12.5\% \\
    \midrule
    \rowcolor{ourshade}\textbf{\method} (progressive) & \textbf{69.66} & \textbf{71.97} & \textbf{58.19} & \textbf{65.21} \\
    Recursive, $\ell=1$ & 67.05 & 69.19 & 57.87 & 64.61 \\
    Recursive, $\ell=2$ & 67.65 & 70.76 & 56.59 & 63.81 \\
    Recursive, $\ell=3$ & 66.15 & 69.15 & 55.37 & 62.26 \\
    \bottomrule
  \end{tabular}%
  }
\end{table}

\section{Additional analyses}
\label{sec:appendix-analyses}

\subsection{Quota effects by image count}
\label{sec:appendix-quota}

Per-image quota gains over global top-$k$ vary with image count on LLaVA-OneVision-7B
at $B=\tfrac1{16}$ (percentage points):
\begin{center}

\resizebox{0.8\columnwidth}{!}{
\begin{tabular}{lccc}
\toprule
 & $N=2$ & $N\in\{3,4\}$ & $N\geq5$ \\
\midrule
Mantis-Eval & $+2.8$ & $+5.6$ & --$^{*}$ \\
MuirBench   & $-1.6$ & $+0.5$ & $+0.7$ \\
\bottomrule
\end{tabular}}

{\footnotesize $^{*}$Mantis-Eval has a single prompt with $N\geq5$, too few to report a gain.}
\end{center}
Quotas improve every reported subset with $N\geq3$ and have mixed effects
at $N=2$ (positive on Mantis-Eval, negative on MuirBench). NLVR2 and
Q-Bench2 are omitted because all their questions have $N=2$; quotas help on
both ($+1.44$ and $+0.90$ points, \cref{tab:ablation-onevision-full}). The benchmark average
in \cref{tab:ablation-onevision-full} therefore hides this variation.

\subsection{Robustness to relevant-image position}
\label{sec:pos_single}

Table~5 summarizes this probe and the joint-evidence probe (\cref{sec:pos_joint}). Both use
LLaVA-OneVision-7B at $B=\tfrac18$. Table~5 takes Mean and Worst from \cref{tab:pos_single}, and
Joint, CF-A, and CF-B from \cref{tab:pos_joint}. Its two \method variants change only the per-image
allocation: \emph{w/ global top-$k$} ranks all visual tokens together, as in Table~6, and
\emph{w/ floor + attention} gives each image the floor $q_{\min}$ and allocates the remainder by
prompt-attention mass (\cref{sec:appendix-abl-extra}).

\noindent\textbf{Protocol.}
We sample 500 MMStar questions approximately uniformly across six categories,
independently of model predictions. Each relevant image is combined with $N-1$ distinct
MMStar distractors for $N\in\{4,8\}$, using nested distractor sets and fixed relative
order. We test every relevant-image position. Only the image number mentioned in
the question changes; the image, options, and answer remain fixed across
positions and methods.
Evaluation uses LLaVA-OneVision-7B at $B=\tfrac18$ (\cref{tab:pos_single}).

\noindent\textbf{Metrics.}
Mean and Worst denote average and minimum accuracy across positions; Gap is the
maximum minus minimum. A small gap alone does not imply high accuracy.

\begin{table}[!htbp]
  \centering
  \caption{\textbf{Relevant-image position robustness} on LLaVA-OneVision-7B at $B=\tfrac18$. Mean/Worst: average/minimum accuracy over relevant-image positions (\%); Gap: best--worst difference (points). Bold marks the best compressed method; Gap is not marked because a small gap alone does not imply high accuracy.}
  \label{tab:pos_single}
  \scriptsize
  \setlength{\tabcolsep}{3.5pt}
  \resizebox{\columnwidth}{!}{%
  \begin{tabular}{lccc|ccc}
    \toprule
    & \multicolumn{3}{c|}{\cellcolor{bbllava}\textbf{$N=4$}} & \multicolumn{3}{c}{\cellcolor{bbllava}\textbf{$N=8$}} \\
    \cmidrule(lr){2-4} \cmidrule(lr){5-7}
    Method & Mean & Worst & Gap$\downarrow$ & Mean & Worst & Gap$\downarrow$ \\
    \midrule
    \textit{Full} & \textit{52.9} & \textit{51.2} & \textit{2.6} & \textit{48.1} & \textit{45.2} & \textit{7.2} \\
    \midrule
    VisionZip~\cite{Yang_2025_CVPR} & 43.1 & 41.2 & 4.6 & 40.1 & 37.8 & 7.2 \\
    FastV~\cite{Chen_2024} & 30.5 & 29.0 & 5.0 & 30.0 & 27.8 & 9.2 \\
    PyramidDrop~\cite{Xing_2025_CVPR} & 28.8 & 27.0 & 3.2 & 28.6 & 26.8 & 3.2 \\
    SparseVLM~\cite{pmlr-v267-zhang25s} & 30.4 & 28.8 & 3.2 & 29.8 & 27.4 & 7.0 \\
    AVTP~\cite{zhang2026multiimagevisualtokenpruning} & 33.5 & 32.0 & 3.0 & 33.5 & 32.0 & 3.6 \\
    CLSE~\cite{chen2026spectralevolutionguidedtokenpruning} & 38.3 & 33.0 & 13.8 & 34.8 & 30.0 & 18.0 \\
    S$^2$Prune~\cite{jia2026s2prunespatiallystructuredvisual} & 44.0 & 43.0 & 2.6 & 40.4 & 38.2 & 5.2 \\
    ToDRE~\cite{li2026todre} & 42.0 & 40.4 & 4.6 & 39.7 & 37.6 & 6.0 \\
    \midrule
    \method\ w/ global top-$k$ & 45.8 & 43.0 & 6.2 & 43.3 & 37.4 & 13.4 \\
    \method\ w/ floor + attention & 45.1 & 41.4 & 9.0 & 43.2 & 36.6 & 14.6 \\
    \rowcolor{ourshade} \textbf{\method} & \textbf{47.6} & \textbf{47.0} & 1.4 & \textbf{45.2} & \textbf{42.2} & 5.4 \\
    \bottomrule
  \end{tabular}%
  }
\end{table}
\subsection{Joint preservation of evidence from two images}
\label{sec:pos_joint}

\noindent\textbf{Protocol.}
We construct 500 image pairs from COCO val2017~\cite{Lin_2014}. Let $I_A$ and $I_B$ denote the
two relevant images. Each pair is associated with a joint question requiring
one fact from each image: 167 \emph{sum}, 167 \emph{comparison}, and 166
\emph{category-matching} questions; for example, ``How many donuts are there
in total across image~$i$ and image~$j$?'', ``Are there more fire hydrants in
image~$i$ than in image~$j$?'', and ``Is the animal in image~$i$ the same kind
of animal as the animal in image~$j$?''. Each fact is also asked separately about its own image, yielding
two single-image (constituent) questions per pair. Pairs are selected
independently of model predictions, and the answers to comparison and matching
questions are balanced between yes and no. For counting questions, each
image contains 1--4 instances of the queried category (crowd annotations
excluded), each covering $\geq0.4\%$ of the image, with pairwise box
$\mathrm{IoU}\leq0.3$. For matching questions, each image contains exactly one
animal category covering $\geq5\%$ of the image.

\noindent\textbf{Placements.}
For $N\in\{2,4,8\}$, we test all $N(N-1)$ ordered positions of $I_A$ and $I_B$
among $N-2$ distractors. The distractors contain no annotated instance of
the queried category (no animal, for matching), and their sets are nested
across $N$ with a fixed relative order. Between placements, only the image
numbers mentioned in the question change. Joint and constituent questions
are separate calls on the same image sequence, with no shared history.

\noindent\textbf{Counterfactual controls.}
We replace $I_A$ (or $I_B$) with a screened image that changes the correct joint
answer, keeping the question, placement, and distractors fixed. Eligible replacements are
available for 394 pairs for $I_A$, 373 for $I_B$, and 315 for both.

\noindent\textbf{Cost and budget.}
Each method processes $158{,}690$ prompts on LLaVA-OneVision-7B with greedy
decoding: $500\times70\times3=105{,}000$ original prompts (70 placements
summed over $N\in\{2,4,8\}$; one joint and two constituent calls) plus
$(394+373)\times70=53{,}690$ replacement prompts. Compressed methods target
$B=\tfrac18$ and realize $12.4$--$12.7\%$ depth-averaged retention.
\Cref{tab:pos_joint} reports $N=4$ and $N=8$. At $N=2$, \method achieves $90.4\%$
joint accuracy, versus $88.1\%$ for \textit{Full} and $79.7\%$ for the best
baseline, VisionZip.

\noindent\textbf{Metrics.}
\emph{Joint} is the joint-question accuracy averaged over pairs and
placements. \emph{Both} requires both constituent answers to be correct;
\emph{All~3} also requires the correct joint answer. \emph{Worst} is the
minimum joint accuracy over placements. \emph{CF-A} (\emph{CF-B}) is the
fraction of eligible pairs whose joint answer is correct both on the
original pair and after replacing $I_A$ ($I_B$).
We use 10{,}000 paired bootstrap resamples over image pairs, retaining all placements,
constituent calls, and replacements per sampled pair.

\begin{table}[!tp]
  \centering
  \caption{\textbf{Joint evidence preservation} on LLaVA-OneVision-7B at $B=\tfrac18$: $500$ COCO pairs over all ordered placements ($12$ for $N=4$, $56$ for $N=8$). Joint: joint-question accuracy; Both: both constituents correct; All 3: all answers correct; Worst: minimum positional joint accuracy. CF-A/CF-B: correct joint answers before and after replacing $I_A$/$I_B$ ($394$/$373$ eligible pairs). Bold marks the best compressed method.}
  \label{tab:pos_joint}
  \scriptsize
  \setlength{\tabcolsep}{1pt}
  \resizebox{1\columnwidth}{!}{%
  \begin{tabular}{lcccccc|cccccc}
    \toprule
    & \multicolumn{6}{c|}{\cellcolor{bbllava}\textbf{$N=4$}} & \multicolumn{6}{c}{\cellcolor{bbllava}\textbf{$N=8$}} \\
    \cmidrule(lr){2-7} \cmidrule(lr){8-13}
    Method & Joint & Both & All 3 & Worst & CF-A & CF-B & Joint & Both & All 3 & Worst & CF-A & CF-B \\
    \midrule
    \textit{Full} & \textit{77.1} & \textit{77.6} & \textit{62.4} & \textit{70.4} & \textit{47.2} & \textit{42.0} & \textit{63.9} & \textit{62.1} & \textit{40.5} & \textit{49.0} & \textit{23.6} & \textit{23.2} \\
    \midrule
    VisionZip~\cite{Yang_2025_CVPR} & 67.2 & 57.7 & 40.9 & 57.6 & 29.4 & 26.1 & 60.7 & 56.7 & 34.9 & 47.6 & 20.5 & 19.8 \\
    FastV~\cite{Chen_2024} & 50.8 & 24.8 & 20.9 & 41.2 & 2.3 & 1.7 & 44.0 & 19.0 & 12.8 & 36.8 & 2.2 & 2.3 \\
    PyramidDrop~\cite{Xing_2025_CVPR} & 49.9 & 24.0 & 16.7 & 44.4 & 1.3 & 1.0 & 42.0 & 17.1 & 8.8 & 36.0 & 2.7 & 3.2 \\
    SparseVLM~\cite{pmlr-v267-zhang25s} & 51.9 & 29.8 & 24.4 & 45.2 & 2.1 & 2.2 & 43.7 & 21.9 & 13.6 & 37.4 & 4.2 & 4.9 \\
    AVTP~\cite{zhang2026multiimagevisualtokenpruning} & 58.0 & 39.4 & 35.2 & 56.4 & 1.9 & 1.6 & 54.6 & 35.9 & 30.5 & 51.2 & 0.7 & 1.0 \\
    CLSE~\cite{chen2026spectralevolutionguidedtokenpruning} & 51.7 & 36.6 & 27.1 & 37.2 & 7.8 & 6.3 & 43.9 & 30.3 & 14.2 & 32.8 & 6.2 & 6.5 \\
    S$^2$Prune~\cite{jia2026s2prunespatiallystructuredvisual} & 64.3 & 52.4 & 38.9 & 58.2 & 23.2 & 22.4 & 54.1 & 44.5 & 23.9 & 42.6 & 13.6 & 12.8 \\
    ToDRE~\cite{li2026todre} & 68.2 & 54.4 & 40.8 & 58.4 & 32.6 & 29.9 & 59.3 & 54.3 & 33.1 & 46.4 & 20.9 & 19.4 \\
    \midrule
    \method\ w/ global top-$k$ & 85.0 & \textbf{83.6} & 73.1 & 81.0 & 60.2 & 51.9 & 74.6 & 74.9 & 57.8 & 64.4 & 42.0 & 37.1 \\
    \method\ w/ floor + attention & 84.0 & 83.1 & 71.9 & 79.6 & 57.4 & 51.1 & 74.6 & 74.9 & 58.0 & 63.2 & 41.5 & 37.4 \\
    \rowcolor{ourshade} \textbf{\method} & \textbf{86.6} & \textbf{83.6} & \textbf{74.7} & \textbf{81.8} & \textbf{62.9} & \textbf{55.0} & \textbf{77.8} & \textbf{77.8} & \textbf{62.7} & \textbf{69.6} & \textbf{47.3} & \textbf{42.5} \\
    \bottomrule
  \end{tabular}%
  }
\end{table}

\clearpage

{
    \small
    \bibliographystyle{content/ieeenat_fullname}
    \bibliography{main}
}

\end{document}